\documentclass[10pt, conference, letterpaper]{IEEEtran}
\IEEEoverridecommandlockouts 

\usepackage{mdframed}
\usepackage{diagbox}  
\usepackage{latexsym,bm}
\usepackage{xcolor}

\usepackage{bbm}
\usepackage{threeparttable}
\usepackage{makecell}
\usepackage[pdftex]{graphicx}
\usepackage{graphicx}
\usepackage{epstopdf}
\usepackage{stfloats}
\usepackage{epsfig}
\usepackage{booktabs}
\usepackage{multirow}
\usepackage[ruled,linesnumbered,noend]{algorithm2e}
\usepackage{amsmath}
\usepackage{amssymb}
 \usepackage{amsfonts}
\usepackage{epstopdf}
\usepackage{subfigure}
\usepackage{graphicx}
\usepackage{array}
 \usepackage{bigstrut}
\usepackage{cite}

\usepackage{latexsym,bm}
\usepackage{xcolor}

\usepackage{makecell}
\usepackage[pdftex]{graphicx}
\usepackage{graphicx}
\usepackage{stfloats}
\usepackage{epsfig}
\usepackage{booktabs}
\usepackage{multirow}
\usepackage[ruled,linesnumbered,noend]{algorithm2e}
\usepackage{amsmath}
\usepackage{amssymb}
\usepackage{amsthm}
\usepackage{subfigure}
\usepackage{graphicx}
\usepackage{array}
\usepackage{comment} 
\usepackage{cite}

\newcommand{\Identity}{{\rm I\kern-.2em l}}
\newcommand{\Expect}{\mathbb{E}}

\newcommand{\subparagraph}{}
 \usepackage{titlesec}
 \titlespacing\section{0pt}{6pt plus 2pt minus 1pt}{4pt plus 2pt minus 1pt}
 \titlespacing\subsection{0pt}{2pt plus 2pt minus 1pt}{2pt plus 1pt minus 1pt}
 \titlespacing\subsubsection{10pt}{2pt plus 0pt minus 2pt}{1pt plus 0pt minus 2pt}
\titleformat{\subsubsection}[runin]{\normalfont\normalsize\itshape}{\arabic{subsubsection})}{5pt}{}[:\,\,]

\usepackage{thmtools}

\declaretheoremstyle[%
  spaceabove=3pt,%
  spacebelow=3pt,%
  headfont=\normalfont\bfseries,%
  bodyfont=\normalfont\itshape,%
  postheadspace=0.5em,%
]{theoremstyle}

\declaretheorem[name={Assumption},style=theoremstyle]{assumption}
\declaretheorem[name={Theorem},style=theoremstyle]{theorem}
\declaretheorem[name={Corollary},style=theoremstyle]{corollary}
\declaretheorem[name={Lemma},style=theoremstyle]{lemma}

\declaretheoremstyle[%
  spaceabove=0pt,%
  spacebelow=3pt,%
  headfont=\normalfont\itshape,%
  postheadspace=1em,%
  qed=\qedsymbol%
]{proofstyle} 

\declaretheorem[name={Proof},style=proofstyle,unnumbered]{prf}

\ifCLASSINFOpdf
 
\else

\fi

\begin{document}
\title{Tackling System and Statistical Heterogeneity for Federated Learning with Adaptive Client Sampling \vspace{-0.05in}
}%

\author{\IEEEauthorblockN{Bing Luo\IEEEauthorrefmark{1}\IEEEauthorrefmark{2}\IEEEauthorrefmark{4},
Wenli Xiao\IEEEauthorrefmark{2}\IEEEauthorrefmark{1}, Shiqiang Wang\IEEEauthorrefmark{3}, Jianwei Huang\IEEEauthorrefmark{2}\IEEEauthorrefmark{1}, 
Leandros Tassiulas\IEEEauthorrefmark{4}}
\IEEEauthorblockA{\IEEEauthorrefmark{1}Shenzhen Institute of Artificial Intelligence and Robotics for Society, China\\
\IEEEauthorrefmark{2}School of Science and Engineering, The Chinese University of Hong Kong, Shenzhen, China\\
\IEEEauthorrefmark{3}IBM T. J. Watson Research Center, Yorktown Heights, NY, USA\\
\IEEEauthorrefmark{4}Department of Electrical Engineering and Institute for Network Science, Yale University, USA\\
Email: \{luobing, lixiang, jianweihuang\}@cuhk.edu.cn,
shiqiang.wang@ieee.org,
leandros.tassiulas@yale.edu}\vspace{-1mm}
\thanks{{The work of Bing Luo %was supported by the AIRS-CUHKSZ-Yale Joint Postdoctoral Fellowship. %The work of Wenli Xiao was supported by Shenzhen Institute of Artificial Intelligence and Robotics for Society.  
Wenli Xiao, and Jianwei Huang were supported
by the Shenzhen Science and Technology Program (JCYJ20210324120011032), Shenzhen Institute of Artificial Intelligence and Robotics for Society, and the Presidential Fund from the Chinese University of Hong Kong, Shenzhen. The work of Leandros
Tassiulas was  supported by
the AI Institute for Edge Computing Leveraging Next Generation Networks
(Athena) under Grant NSF CNS-2112562 and Grant NRL N00173-21-1-G006. (Corresponding author: Jianwei Huang.)}}
\vspace{-0.3in}
}

\maketitle

\begin{abstract}
%It is well understood that the effectiveness and efficiency of 
%It is well understood that 
Federated learning (FL) algorithms usually  sample a fraction of clients  in each round (partial participation) when the number of participants is large and the server's communication bandwidth is limited. Recent works on the convergence analysis of FL have focused on unbiased client sampling, %due to massive participants and limited system bandwidth. 
e.g., sampling uniformly at random, % or in proportional to their data size, 
{which suffers from slow wall-clock time for convergence  %with respect to wall-clock time 
due to high degrees of system heterogeneity %(i.e., diverse communication delays) 
and statistical heterogeneity.} %(i.e., unbalanced and non-i.i.d. data).  
 %. The former, due to different computational and communication capabilities, results in different local communication delay  
This paper aims to design an adaptive client sampling algorithm that tackles both system and statistical heterogeneity to minimize the wall-clock convergence time. % with a convergence guarantee. %total learning time while ensuring convergence. % (instead of the number of training rounds). 
%minimize the wall-clock time  while ensuring convergence. 
%Theoretically,
We obtain a new tractable convergence bound for FL algorithms with arbitrary client sampling probabilities. %that can adapt to arbitrary client  sampling probability. %via careful sampling and aggregation design.
Based on the bound, %we successfully formulated an analytical optimization problem. %
we analytically establish the  relationship between the total learning time and sampling probabilities, which results in  a non-convex optimization problem for training time minimization. %and client sampling probability. % with a new convergence bound, %which adapts to partial participation and different number of local updates. % and  different participation rate.
%allow partial clients participation and perform more local model updates
%Our convergence bound explicitly shows that clients with larger datasize and local gradient norm should be sampled with higher probability for speeding up the convergence rate with respective of the number of communication rounds.
%Particularly, 
%To efficiently solve the time-wise %wall-clock time 
%minimization problem, To achieve the global optimal solution, 
We design an efficient algorithm % substitute sampling method 
for learning the unknown parameters in the convergence bound and develop a low-complexity algorithm to approximately solve the non-convex problem. % with low complexity.  %with  marginal overhead. %develop a substituted sampling based algorithm to learn the convergence-related unknown parameters  with marginal overhead. 
%Our solution reveals insightful sampling principles %that higher sampling %the optimal sampling strategy %for speeding up FL should give higher 
%probability should give to clients who have faster communication time and larger product value of data proportion and gradient norm,  %norm should be sampled with high probability, %the former reduces the per-round time, the later reduces the variance and the expected round number, and the optimal sampling strategy achieves the balance between system and statistical heterogeneity.
%which identify the impact and interplay %impact and trade-off 
%between system and statistical heterogeneity. % in speeding up FL. % in terms of convergence time.
%Our adaptive sampling convergence bound but also allows for standard FL settings with multiple local updates and partial participation. 
%we obtained a new convergence bound that established the relationship between the expected number of  communication rounds and the sampling probability with
%Empirically, 
 Experimental results from both hardware prototype and  simulation demonstrate that  
 %we evaluate our theoretical results  both in a simulated environment and on a hardware prototype.
 %Experiments show that, 
 our proposed sampling scheme
 significantly reduces the convergence time % reaches the target loss and accuracy with significantly less time 
 compared to several baseline sampling schemes. 
 %consistently achieves lower training loss and higher test accuracy  than baseline sampling  schemes for given time budgets. 
 Notably, our scheme in hardware prototype  %with EMNIST dataset
 spends $73$\% less time than the  uniform sampling baseline for reaching the same target loss.
%demonstrates that  our scheme   reaches the target loss up to 4$\times$ faster than baseline uniform sampling. %,the superiority of our sampling scheme in terms of convergence  has a superior wall-clock time performance converges with , e.g.,  for various datasets and  heterogeneous system and statistical settings. compared to 
%for a fixed wall-clock budget, our method provide a reduction of the train losses of up to an order of magnitude and a relative improvement of test error Y\%. Comparing to uniform at random sampling, we show that our method consistently achieve lower training loss and test error for equalized wall-clock time.} %parameters%with  optimality error less than 5\% in various FL settings. %with different datasets, different  optimization  goals,  and  heterogeneous  system parameters. %Particularly, we show that the well-known  communication-efficient FL design, on the contrary, might %severely slow down the learning. %
%cause a considerably long learning time. %$\gamma$. %, which reduces the total cost significantly.
%which requires practitioners careful design
%\\which can exacerbate the adverse effects of data heterogeneity \\
%true speed of error convergence with respect to time instead of the number of iterations
\end{abstract}

\IEEEpeerreviewmaketitle

\section{Introduction}
\vspace{-1mm}

%To tackle this challenge, 
{Federated learning (FL)} %has recently emerged as an attractive distributed machine learning (DML) paradigm, which 
enables many clients\footnote{%Depending on the type of clients, FL can be categorized into cross-device FL and cross-silo FL (clients are companies or organizations, etc.) \cite{kairouz2019advances}. 
We %study cross-device FL and 
use ``device'' and ``client'' interchangeably in this paper.}
 to collaboratively train a  model under the coordination of a central server while keeping the training data decentralized and private (e.g., \cite{kairouz2019advances,yang2019federated,mcmahan2017communication}). %
%Similar to conventional DML systems, FL let the clients  perform most of the computation and a server iteratively aggregate their computed updates. 
Compared to traditional distributed machine learning  techniques,   
FL has two unique features (e.g., \cite{bonawitz2019towards,li2020federated,li2018federated,yu2018parallel,yu2019linear,wang2018adaptive}), as shown in Fig.~1.  {First, 
%The first and most widely used FL algorithm is Federated Averaging (FedAvg), which performs $E$ steps%\footnote{It is known that when using $E=1$ and full batch, FedAvg is equivalent to full gradient decent which normally requires a large number of communication rounds to reach a target performance.}
%of stochastic gradient decent (SGD) in parallel on $K$ (i.e., a fraction of) clients before aggregating their updated models \cite{hbm}. 
 %two key challenges distinguish federated learning from standard DML: high degrees of \emph{statistical and systems heterogeneity} \cite{li2018federated, mcmahan2017communication} %In FL settings, 
clients are massively distributed and with diverse and low communication rates (known as \emph{system heterogeneity}), where stragglers can slow down the physical training time.\footnote{As suggested \cite{bonawitz2016practical,avent2017blender,konevcny2016federated,9567711,sun2021pain}, we consider mainstream synchronized FL in this paper %as It was suggested that  FL algorithms should operate synchronously 
due to its composability with other
techniques (such as secure aggregation protocols and differential privacy).}} %, and model compression. 
%For synchronous FL the round time is limited by the slowest  client (straggler).} 
{Second, the training data are distributed 
in a {non-i.i.d.} and {unbalanced} fashion across the clients  (known as \emph{statistical heterogeneity}), which negatively  affects the convergence behavior.}% even with full client participation \cite{yu2018parallel,yu2019linear,wang2018adaptive}.}

Due to limited communication bandwidth and across geographically dispersed devices, FL algorithms (e.g., the de facto FedAvg algorithm in \cite{mcmahan2017communication}) usually %As a de facto optimization algorithm in the FL community, federated averaging (FedAvg) 
perform {multiple local iterations} on \emph{a fraction of randomly sampled clients (known as partial participation)}  and then aggregates their resulting local model updates via the central server periodically \cite{mcmahan2017communication, bonawitz2019towards,li2018federated,li2020federated}. 
%In this way, each client only sends its locally computed model update to the server but is able to learn the shared model from other participants, and by performing multiple local iterations, the communication burden between the central server and clients can be significantly alleviated. %
%In particular, the most popular and de facto optimization algorithm is federated averaging (FedAvg),  %is a communication-efficient algorithm that has emerged as the de facto optimization method in the federated setting. At each iteration, FedAvg first locally 
%which performs multiple local stochastic gradient descent (SGD) in parallel on a fraction of randomly selected devices. %, and then aggregates the resulting local model updates via the central server periodically.  %where E is a small constant and K is a small fraction of the total devices in the network. The devices then communicate their model updates to a central server, where they are averaged.
%FedAvg
Recent works have provided theoretical convergence analysis that demonstrates the effectiveness of FL with partial participation    %FL has demonstrated empirical success and theoretical convergence guarantees in recent works
in various non-i.i.d. settings %for both convex or non-convex settings %various statistical heterogeneous settings, e.g., unbalanced and non-i.i.d. data distribution 
\cite{%li2018federated,  
haddadpour2019convergence,karimireddy2019scaffold,yang2021achieving,li2019convergence, qu2020federated}. %and established a convergence guarantee for strongly convex problems. 
%, it does not fully address the underlying challenges associated with heterogeneity
\begin{figure}[!t]
	\centering
	\includegraphics[width=8.8cm,height=5.4cm]{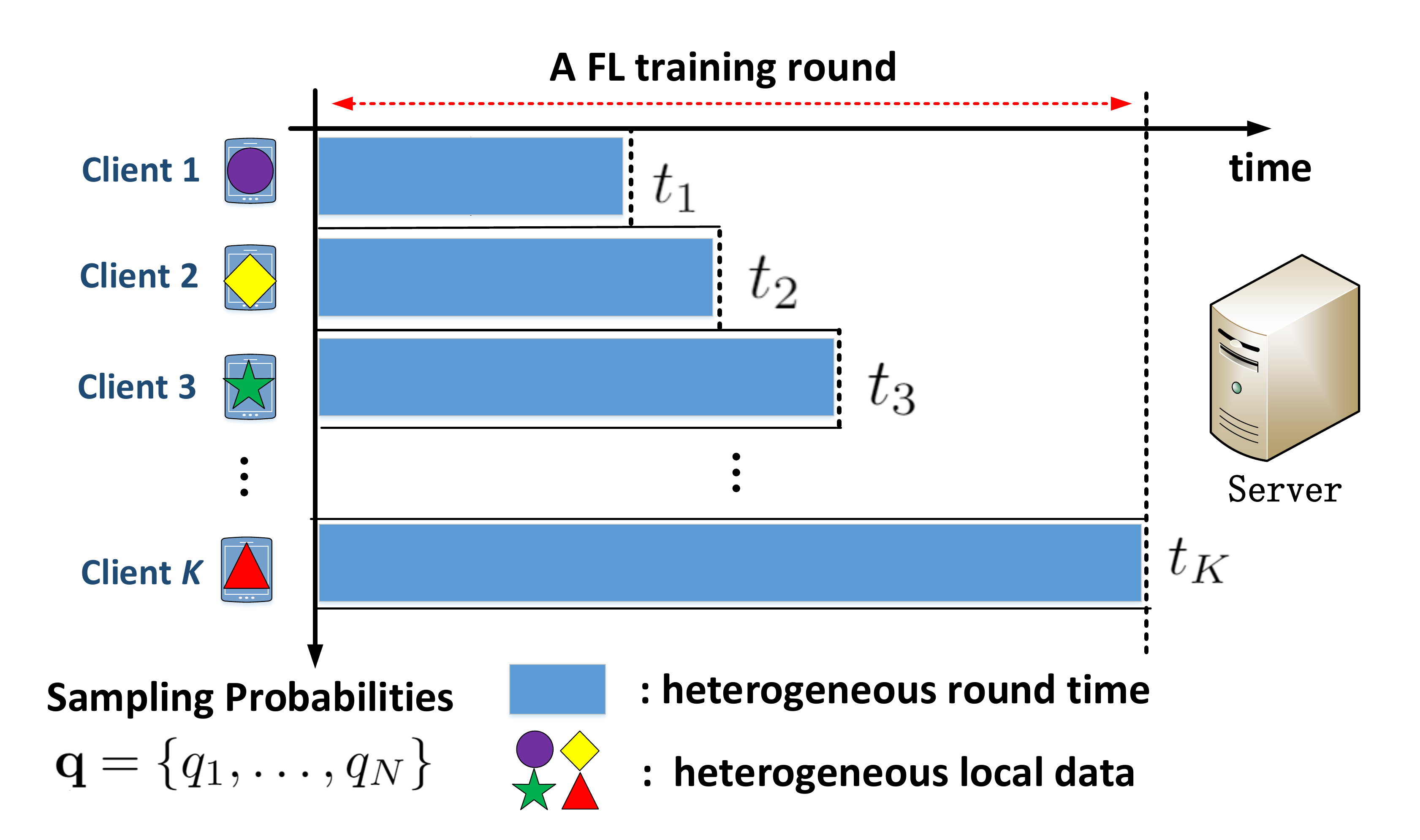}
% 	\vspace{-2mm}
	\caption{An FL training round with system and statistical heterogeneity,  with $K$ out of $N$ clients sampled  according to probability  $\mathbf{q}=\{q_1, \ldots, q_N\}$.}%Heterogeneous FL system: devices with variant computation and communication capacities.}
	\label{fig:intro}
% \vspace{-3mm}
\end{figure}

However, %the partial participated clients these convergence-guaranteed works  
these prior works \cite{%li2018federated,  
haddadpour2019convergence,karimireddy2019scaffold,yang2021achieving,li2019convergence, qu2020federated} have focused on sampling schemes that select clients   
%the client sampling schemes in these convergence-guaranteed works are 
%restricted to be %\emph{unbiased}, e.g.,  
uniformly at random or proportional to the clients' data sizes, %such that the aggregated model updates are unbiased towards that with full client participation, 
{which often suffer from slow %error
convergence with respect to wall-clock (physical) time\footnote{We use wall-clock time to distinguish from the number of training rounds. %adopted in many communication efficient FL works.
} due to high degrees  of the system  and  statistical  heterogeneity}. %(instead of the number of training rounds). %yields a longer %naturally slows down  %causes slower
 %speed with respect to .%\emph{wall-clock time}. %due to {statistical} and {system heterogeneity}. 
This is because the total FL time {depends on \emph{both the number of training rounds} for reaching the target precision and %error-convergence rate with the number of training round and 
\emph{the physical time in each round}} \cite{stragglers}. 
%For unbiased sampling, clients 
%Specifically,
%\begin{itemize} \item 
Although uniform sampling  guarantees that the aggregated model update in each round is unbiased towards that with full client participation, the aggregated model may have a high variance %compared to that of full participation 
due to data heterogeneity, thus,  \emph{requiring more training rounds to converge to a target precision}. % which negatively affects the convergence rate, i.e., \emph{requiring more training rounds}. %to achieve a target precision. %In other words, clients with good and poor data quality are sampled equally.
Moreover, considering clients' heterogeneous communication delay, uniform sampling also suffers from the straggling effect, as the probability of sampling a straggler within the sampled subset in each round can be relatively  high,\footnote{For example, suppose there are 100 clients with only 5 stragglers, then the probability of sampling at least a straggler for uniformly sampling 10 clients in each round is more than 40\%.} thus \emph{yielding a long per-round time}.
 
%In a nutshell, clients with valuable data may have poor communication rate, whereas clients who can feedback their updates fast may have low-quality data and affect the convergence. %importance sampling only samples clients with good quality, which reduces the required round number, but these clients may have a poor communication spped, which reduces a longer wall-clock round time if sampled with high probabilty.%In other words, clients with good or poor data quality, and with fast or slow communication time are all sampled with  equal probability.

%Recent FL client sampling works 
One effective way of speeding up the convergence with respect to the number of training rounds is  %the number of training rounds, 
%recent works proposed %\emph{importance sampling} %, where clients are sampled according to 
to choose clients according to some sampling distribution where ``important" clients have  high probabilities \cite{zhao2015stochastic,needell2014stochastic,alain2015variance,gopal2016adaptive}. For example, recent works adopted  {importance sampling} approaches based on %schemes via exploring 
clients' statistical property %, e.g., data size, loss or gradient information 
\cite{chen2020optimal,rizk2020federated,nguyen2020fast,cho2020client}.
%An effective way of speeding up FL is to design %biased 
%non-uniform sampling, where clients are sampled according to some probability distribution $\mathbf{q}$ with ``important" clients having  higher probabilities. 
%via selecting valuable clients with higher probability. 
%some recent works adopted \emph{importance sampling} approach %to select clients 
%via exploring the statistical property, e.g., sampling clients based on data quantity (data size), or data quality (local loss or gradient norm), to {speed up the error-convergence rate with respect to the number of iterations}. % by reducing the training rounds, however, without considering heterogeneous training time [a-b]. %,
However, their sampling schemes %a limitation of these works is that they
did not account for the heterogeneous physical time  in each round,  especially under  straggling circumstances. % for convergence guarantees [a-b]. 
%improve the error convergence rate with respect to required training rounds improve the error convergence rate with   
Another line of works aims to %mitigate stragglers, and thus 
minimize the learning time via optimizing client selection and scheduling based on their heterogeneous system resources \cite{tran2019federated,chen2020convergence,9207871,stragglers,nishio2019client,chai2020tifl, yang2019scheduling,jin2020resource,9579038,wang2020optimizing,wang2019adaptive,luo2020cost,tu2020network,wang2021device}. %to mitigate stragglers
% or using online learning or reinforcement learning based client selection methods \cite{jin2020resource,han2020adaptive,wang2020optimizing, xia2020multi}. %optimize the FL learning time by 
%designed client selection and scheduling schemes based on their heterogeneous resources to mitigate stragglers, which,
However,  {their optimization schemes did not consider
% optimize %the impact of clients' heterogeneous data on the FL algorithm
% the internal procedure of the  %optimization of the 
% FL algorithm for clients' heterogeneous data, e.g., 
how client selection schemes influence the convergence behavior due to data heterogeneity and thus may negatively affect the total learning time.}%, such as the impact of data quantity or quality. %, which does not consider controlling the internal procedure of the FL algorithm itself., 

In a nutshell, the fundamental limitation of existing works is the \emph{lack of joint consideration of the impact of the inherent system heterogeneity and statistical heterogeneity on client sampling}. %the fundamental limitation of existing works is the {\emph{lack of consideration of how client sampling affects both system and statistical heterogeneity}}. 
In other words,  clients with {valuable data} may have poor communication capabilities, whereas those who communicate fast may have low-quality data.
% e.g.,  clients with valuable data may have poor communication rate, whereas those who communicate fast may have low-quality data. % and affect the convergence. 
%However, %\emph{none of these schemes obtain the optimal sampling strategy that jointly addresses systems and statistical heterogeneity with convergence guarantees that adapts to variant participation rate and multiple local updates.}
%\emph{none of these schemes obtained the optimal sampling strategy that adapts to both systems and statistical heterogeneity.}  %while ensuring 
% with convergence guarantees for variant participation rate and multiple local updates.} %These observations motivate the following key question:\\
This motivates us to study the following key question. 

%Motivated by these observations, the key question we ask is that: 

%\setlength{\lineskip}{5em} 
{\setlength{\parskip}{0.3em} \noindent\textbf{Key Question:} \emph{How to design an optimal client sampling scheme that tackles both system and statistical heterogeneity to achieve  fast convergence with  respect to wall-clock time?}} %to speed up FL convergence time.}%with respect to wall-clock time?} %
%\vspace{0.3mm}

{\setlength{\parskip}{0.3em}The challenge of this question is threefold: (1)  It is difficult to obtain an analytical FL convergence result for arbitrary client sampling probabilities. {(2)} The  total learning time minimization problem can be complex and non-convex %as the objective function would be  nonlinear polynomial in sampling probability
due to the straggling effect. % with client sampling. %, and even an approximate formulation can be non-convex to solve. %It is challenging to establish an analytical relationship between the total wall-clock time and sampling probability with the coupled convergence constraint due to straggling effect. 
{(3)} {The optimal client sampling solution contains unknown parameters from the convergence result,} which we can only estimate during the learning process (known as the \emph{chicken-and-egg problem}).} 
%represent how client's sampling probability affects the wall-clock convergence behavior that captures both system and statistical properties of FL; %optimization 
%optimal sampling probability 
%requires to handle the \emph{chicken-and-egg problem}  as the convergence-related parameter values that lead to the optimal sampling solution can only be known during the learning process. 

In light of the above discussion, we
state the main results and key contributions of this paper as follows:

%1. what is the optimal client sampling probability that  
%Therefore, the question we asked 
%an efficient FL client sampling strategy should take into account and balance both systems and statistical heterogeneity to speed up the training process. 

%The heterogeneity caused by imbalanced computational capabilities (Ignatov et al., 2019) and communication bandwidth may significantly degrade the training speed. The global training process is limited by the slower clients, i.e., stragglers. If faster clients wait for slower ones, the overall training speed will become slow. Otherwise, the data in slower clients cannot be well utilized for global model learning

%This is because the convergence speed of FL depends on two factors: 1) the error in the trained model versus the number of iterations, and 2) the number of iterations completed per second. Traditional single-node SGD analysis focuses on optimizing the first factor, because the second factor is generally a constant when SGD is run on a single dedicated server

%For theoretical analysis, 
\begin{itemize}
\item \emph{Optimal Client Sampling for Heterogeneous FL:}   We study how to design the optimal client sampling strategy to
minimize FL wall-clock time  with convergence guarantees. To the best of our knowledge, this is the first work that aims to optimize client sampling probabilities to address both system and statistical heterogeneity.  

%minimize the expected total wall-clock training time while ensuring the convergence. specifically, client sampling probability,
%speed up FL error convergence with respect to wall-clock training time. %We show how client sampling affects the   
%original, non-convex, after 

    \item  \emph{Convergence Bound for Arbitrary Sampling:} %To establish %connect %establish the  relationship %with the expected total wall-clock time. %    between 
    %the analytically relationship between the  wall-clock time and sampling probability with the coupled convergence constraint, 
   Using an adaptive client sampling and model aggregation design, we obtain a new tractable convergence upper bound for FL algorithms with arbitrary client sampling probabilities. % for FL algorithms with partial participation and multiple local iterations. %The intuition for the convergence guarantee is that if we biased sampled some clients, we need to inversely weighted their updates, so that the  aggregated global model can be unbiased towards that with full participation.  %. As far as we know, this is the first analytical convergence bound that adapts client sampling probability  
    %for standard FL algorithms with partial client participation and multiple local iterations. %Moreover, our bound also identifies how data quantity and data quality affect the convergence performance.  %As far as we know, this is the first convergence result also reals the impact of  identifies the impact of data quantity and data quality on client sampling. 
 %   Based on the derived convergence bound, we  successfully establish %connect %establish the  relationship %with the expected total wall-clock time. %    between 
   This enables us to establish the analytical relationship between the total learning time and client sampling probabilities and formulate a non-convex training time minimization problem.% constraint.
    
%  \item {Optimal Sampling Properties:}
  
    \item \emph{Optimization Algorithm and %Solution Properties% Design  
    Sampling Principle:}   We propose a low-cost substitute sampling approach to learn the convergence-related unknown parameters and develop an efficient algorithm to approximately solve the non-convex problem with low computational complexity. %substitute sampling based algorithm % low-complexity algorithm     to learn the convergence-related unknown parameters with marginal   overhead. % and then achieve the global optimum.  % that lead to the optimal sampling probability. %     with marginal estimation overhead.
    {Our solution characterizes the impact of %and interplay between 
    communication time (system heterogeneity) and data quantity and quality (statistical heterogeneity) on the optimal client sampling design.}
         %Our solution reveals insightful sampling principles, which  analytically identifies the  impact and interplay between system  and  statistical  heterogeneity for the first time. Notably, we show that high sampling probability should be allocated to clients who communicate fast and have large product value of datasize (data quantity) and gradient norm (data quality).} %, which, for the first time, analytically identifies the  impact and  interplay between system  and  statistical  heterogeneity.  % for speeding up FL training.

    \item \emph{Simulation and Prototype Experimentation:} We evaluate the performance of our proposed algorithms through both  %via extensive experiments 
  %  with real and synthetic datasets, both 
  a simulated environment and a hardware prototype. % with Raspberry Pi and Jetson Nano devices. %with 20 Raspberry Pis serving as devices and a laptop serving as the central server. 
Experimental results demonstrate that for both convex and
non-convex learning models, %compared to baseline uniform sampling,  %our method provide a reduction of the train losses of up to an order of magnitude and a relative improvement of test error Y\%. Comparing to uniform at random sampling, 
our proposed sampling scheme significantly  reduces  the  convergence  time %reaches the target loss and accuracy with significantly less time 
compared to several baseline sampling schemes. %our method consistently achieves lower training loss and higher test accuracy than baseline sampling schemes for  given wall-clock budgets. %parameters%with  optimality error less than 5\% in various FL settings. %with different datasets, different  optimization  goals,  and  heterogeneous  system parameters.
For example, with  our hardware prototype and the EMNIST dataset,  our sampling scheme spends $73$\% less time than baseline uniform sampling for reaching the same target loss.
%converges to the target loss up to 4$\times$ faster than baseline uniform sampling.
%Moreover, empirical results demonstrate the results derived from homogeneous system achieves near-optimal performance in heterogeneous systems. %Moreover, although our theoretical analysis holds for convex loss function, rigorous experiments on training non-convex deep convolutional neural networks show that our derived design principles are still valid.    %We also illustrate the interplay between minimizing learning time and energy consumption and show how the weighting price factor balances the trade-off between the two objectives.
\end{itemize}

%(e.g., converge to a predefined accuracy level)

%Prefernces needs, prices for energy consumption and price for time

%on-device federated learning where distributed mobile user interactions are involved and communication cost in massive distribution

%Unlike centralized ML, energy consumption is important for 

%resource constraints, it is necessary to optimize the energy efficiency for FL
%implementation.

%Nevertheless, FL has three unique characters that distinguish
%it from the standard parallel optimization \cite{Tian}

%In an attempt to tackle high communication cost, optimization methods that allow for local updating and low participation are a popular approach for federated learning  \cite{hbm, smith}

%For FL settings, the data are massively distributed over a large number of devices, and the connection between the central server and a device is usually slow and unstable, which motivates communication-efficient FL algorithms (McMahan et al., 2017;
%Smith et al., 2017; Sahu et al., 2018; Sattler et al., 2019). Federated Averaging (FedAvg) is the first 

%A careful system design for cost minimization is crucial in resource constrained systems. 

\section{Related Work}
Active client sampling and selection play a crucial role in addressing the statistical and system heterogeneity challenges in cross-device FL. 
In the existing literature, the research efforts in speeding up the training process mainly focus on two aspects:  %Existing literature mainly proposed 
importance sampling and resource-aware optimization-based approaches.% to speed up the training process, which we describe as follows.  

%convergence-guaranteed works mostly assume uniform client sampling, such that the aggregated model update is unbiased towards that with full client participation \cite{li2018federated, haddadpour2019convergence,karimireddy2019scaffold,yang2021achieving,li2019convergence, qu2020federated}. However, the learning time can be adversely affected due to system and statistical heterogeneity. Our work along with the following work

define
The goal of importance sampling is %was first motivated 
to reduce the variance in traditional optimization algorithms based on  stochastic  gradient  descent (SGD), where SGD draws data samples uniformly at random during the  learning process (e.g., \cite{zhao2015stochastic,needell2014stochastic,alain2015variance,gopal2016adaptive}). 
Recent works have adopted this idea in FL systems to improve communication efficiency via designing  client sampling strategy. %reducing the variance of the aggregated model update. %,  and thus improve communication-efficiency 
 %Similar idea has 
%The intuition of IS is that we can estimate a random variable by seeing important examples more often, but use them less. Similay
%and has been studied in FL
%client sampling has also been adopted in federated learning (FL) systems for improving communication-efficiency \cite{chen2020optimal,nguyen2020fast,rizk2020federated,cho2020client}. The motivation of these works is that most convergence-guaranteed FL algorithms require clients to be sampled  either uniformly at random or proportional to their local dataset size \cite{li2018federated,  haddadpour2019convergence,karimireddy2019scaffold,yang2021achieving,li2019convergence, qu2020federated}, which leads to a high-variance aggregated update and slows down the convergence due to data heterogeneity. 
Specifically, clients with ``important" data would have higher probabilities to be sampled  in each round. For example, existing works use clients' local gradient information (e.g., \cite{chen2020optimal,rizk2020federated,nguyen2020fast}) or local losses (e.g., \cite{cho2020client}) to measure the importance of clients' data.
%The data importance is often measured by clients' gradient norm \cite{chen2020optimal,rizk2020federated,nguyen2020fast}, or their local losses \cite{cho2020client}.  %considered an asynchronous FL fashion and adopt IS for both client sampling and the data point sampling within each client.\cite{cho2020client} showed that the importance clients can be measured by their local losses. 
However, these schemes did not consider the  speed of error convergence with respect to \emph{wall-clock time}, especially the straggling effect due to heterogeneous transmission delays. %,they did not consider optimization for cost/resource efficiency 
%\ \cite{cho2020client} and biased selection selecting clients with

Another line of works aims to minimize wall-clock time via resource-aware optimization-based approaches, % \cite{  chen2020convergence,tran2019federated,shi2020device,nishio2019client,chai2020tifl, wadu2020federated, yang2019scheduling,tu2020network,wang2021device}.  Specifically, these works considered 
such as CPU frequency allocation (e.g., \cite{tran2019federated}),  and communication bandwidth allocation   (e.g., \cite{chen2020convergence,9207871}), straggler-aware client scheduling    (e.g., \cite{stragglers,nishio2019client,chai2020tifl, yang2019scheduling,jin2020resource,wang2020optimizing,9579038}), parameters control (e.g., \cite{9579038,wang2020optimizing,wang2019adaptive,luo2020cost}), and task offloading (e.g., \cite{tu2020network,wang2021device}). 
While these papers provided some novel insights,  %their optimization schemes  did  not consider  the  impact  of  the  internal  FL  algorithm  itself  forclients’ heterogeneous data.
their optimization approaches %client selection and scheduling 
 %in these works 
did not consider how client sampling affects the total wall-clock time and thus are orthogonal to our work.%their schemes affect the convergence behavior due to data heterogeneity and thus affect the total learning time.

Unlike all the above-mentioned works, our work focuses on how to design the %the client sampling probability %
optimal client sampling strategy that 
tackles both system and statistical heterogeneity to minimize the  wall-clock  time  with  convergence guarantees. %Our work is mostly related to \cite{stragglers}, where the authors proposed a straggler-resilient client selection scheme %to speed up the wall-clock time
%via leveraging the interplay between system and statistical heterogeneity.
%both the clients' analyzed the interplay between statistical and system heterogeneity. 
%However, the idea of their approach is to reduce the overall run-time via sampling faster clients first and gradually involve the slower nodes until full participation, which is  inherently different from our sampling with  partial participation. %because the number of sampled clients in their scheme increased round by round. % in they require the full clients participation at the end of did not optimize  client sampling probability for heterogeneous data. %The later work, however, %is different from our work because their client sampling 
%aims to optimize the error convergence with respect to communication rounds instead of wall-clock time, and their convergence is based on \emph{asychronized} FedProx algorithm from \cite{li2018federated}, while our sampling and convergence result are valid for main stream synchronized FL algorithms, e.g., FedAvg. %Moreover, their work did not model the relationship between the total learning time and their sampling scheme, whereas our work aims to establish the analytical relationship between the two and achieve the optimal sampling probability via solving an optimization problem. 
%did not model the total learning time and did not optimze  how their sampling affects the total learning time consider ,  show analyze the straggling effect in their sampling and did not show how their sampling affects the total wall-clock time.
In addition, most existing works on FL are based on computer  simulations. In contrast, we implement our algorithm in an actual hardware prototype with resource-constrained devices, which allows us to capture real system operations. % with variance. % in a configurable WiFi network. 

%The active sampling approach is another direction in which the server aims for aggregating as many local updates as possible within a predefined time span (Nishio and Yonetani, 2019). More recently, Wang et al. (2020) proposed a normalized averaging method to mitigate stragglers in federated systems and the objective inconsistency due to mismatch in clients’ local updates. Deadline-based computation has also been proposed to mitigate stragglers in decentralized training (Reisizadeh et al., 2019b). In a different yet related direction, various federated learning algorithms have been studied to address the heterogeneity in clients’ data distributions (Karimireddy et al., 2019; Haddadpour et al., 2020; Li et al., 2018; Reisizadehet al., 2020a; Mohri et al., 2019; Reddi et al., 2020).

The organization of the rest of the paper is as follows.  Section~\ref{sec:systemModel} introduces the system model and problem formulation. Section~\ref{sec:convergence} presents our new error-convergence  bound  with arbitrary client sampling.   Section~\ref{sec:optimizationProblem} gives the  optimal client sampling algorithm and solution insights.  %We provide theoretical analysis on the solution properties in Section~\ref{sec:property}.
 Section~\ref{sec:experimentation} provides the simulation and prototype experimental results. We conclude this paper in Section~\ref{sec:conclusion}.

%this is becasue unlike traditional datacenter based ML, the computation and communications in FL consumes a large amount of energy from devices, and mostly from their limited battary, especially for IoT devices or remote sensor nodes. 

%The research scope on FL can be divided into four areas: improving the efficiency and effectiveness, privacy protections, robustness to attacks and failures, and addressing bias and ensuring fairness \cite{peter}. This work is focused the first issue.  

%\newpage

\section{%Modeling the Cost Minimization Problem  FL 
Preliminaries and System Model}
\label{sec:systemModel}

We start by  summarizing the basics of FL and its de facto algorithm FedAvg with unbiased client sampling. Then, we introduce the proposed adaptive client sampling for statistical and system heterogeneity based on FedAvg.  Finally, we present our formulated optimization problem. 

\subsection{Federated Learning (FL)}% and Federated Averaging Algorithm}
%Consider a scenario with a large population of mobile clients that have data for training a machine learning  model. %that they want to maintain as secret. 
%Due to privacy and bandwidth limitation issues, it is not desirable for clients to send their raw %large amount of 
%data to a high-performance data center. %FL is a decentralized learning framework that aims to resolve this problem. %enables a large amount of edge devices to coordinately learn a shared model without data sharing.
%Suppose % a scenario with
%a large number of $N$ clients want to train a machine learning  model. Due to privacy and bandwidth limitation concerns, it is not desirable for them to send the raw data to a high-performance data center. FL is a decentralized learning framework that aims to resolve this problem.

Consider a federated learning system involving a set of $\mathcal{N}={1,\ldots, N}$ clients, coordinated by a central server. Each client $i$ has $n_i$ local training data samples ($\mathbf{x}_{i, 1}, \ldots, \mathbf{x}_{i, n_{i}}$), and the total number of training data across $N$ devices is $n_\textnormal{tot} :=\sum\nolimits_{i = 1}^N n_i$. 
Further, define $f(\cdot,\cdot)$ as a loss function where ${{f}\left( \mathbf{w}; \mathbf{x}_{i, j} \right)}$ indicates how the machine learning model parameter $\mathbf{w}$ performs on the input data sample $\mathbf{x}_{i, j}$. Thus, the local loss function of client $i$ can be defined as 
\begin{equation}
\label{lo_ob}
{F_i}\left( \mathbf{w} \right) := \frac{1}{{{n_i}}}\sum\nolimits_{j =1}^{n_i} {{f}\left( \mathbf{w}; \mathbf{x}_{i, j} \right)}.
\end{equation}
Denote $p_i=\frac{n_i}{n_\textnormal{tot}}$ as the weight of the $i$-th device such that $\sum\nolimits_{i = 1}^N p_i=1$. Then, by denoting $F\left( \mathbf{w} \right)$ as the global loss function, the goal of FL is to solve the following optimization problem 
\cite{kairouz2019advances}:
%Mathematically, the goal of FL is to solve the following optimization problem in a distributed fashion \cite{kairouz2019advances,mcmahan2017communication}:
\begin{equation}
\label{gl_ob}
\min_{\mathbf{w}}  F\left( \mathbf{w} \right) :=\sum\nolimits_{i = 1}^N{p_i}{F_i}\left( \mathbf{w} \right).
\end{equation}
%where the objective $F\left( \mathbf{w} \right)$ is the global loss function, %and $\mathbf{w}$ is the model parameter vector. %, $N$ is the total number of devices, and $p_k$ is the weight of the $i$-th device such that $\sum\nolimits_{i = 1}^N p_i=1$.  Suppose the $k$-th device has $n_k$ training data samples ($\mathbf{x}_{i, 1}, \cdots, \mathbf{x}_{i, n_{i}}$), and the total number of training data samples across $N$ devices is $n :=\sum\nolimits_{i \!=\! 1}^N n_i$, then we have $p_i=\frac{n_k}{n}$.
%In general, the global objective measures the local empirical risk %$F\left( \mathbf{w} \right)$ is global objective function with  model parameter $\mathbf{w}$, , and  The local distribution $\mathcal{D}_k$. 
%   ${F_i}\left( \mathbf{w} \right)$ is the local loss function of client $i$, which is defined as

%where $f(\cdot)$ represents a per-sample loss function, %e.g., mean square error and cross entropy applied to the output of a model
%with parameter $\mathbf{w}$ and input data sample $\mathbf{x}_{i, j}$. %For clients with non-i.i.d. training data, the minimum values $F^*$ and $F_i^*$ for client $i$ are usually different. %$\mathbf{x}_{i,j}$.% \cite{wang2019adaptive}. 

The most popular and de facto optimization algorithm to solve \eqref{gl_ob} is FedAvg \cite{mcmahan2017communication}. Here, denoting $r$ as the index of an FL \textit{round}, %the number of communication rounds, % between the server and clients, 
we  describe one round (e.g., the $r$-th) of the FedAvg algorithm as follows: 
\begin{enumerate}
    \item The server \emph{uniformly at random samples} a subset of  $K$  clients (i.e., $K\! :=\! \left| \mathcal{K}^{r}\right|$ with $\mathcal{K}^{r}\! \subseteq\!  \mathcal{N}$) and {broadcasts} the latest model $\mathbf{w}^r$ to the selected clients. 
    \item Each sampled client $i$ chooses $\mathbf{w}_i^{r,0}\!=\!\mathbf{w}^r$, and runs $E$ steps\footnote{$E$ is originally defined as epochs of SGD in \cite{mcmahan2017communication}. In this paper, we denote $E$ as the number of local iterations for theoretical analysis.} {of local SGD} on~\eqref{lo_ob} to compute an updated model $\mathbf{w}_i^{r,E}$. Then, the sampled client lets $\mathbf{w}_i^{r+1}\!=\!\mathbf{w}_i^{r,E}$ and send it back to the server.
     
    \item  The server \emph{aggregates} (with weight $p_i$) the clients' updated model  and computes a new global model $\mathbf{w}^{\tau+1}$.% for the next round.
\end{enumerate}
%First, %in each {round}~$r$, FedAvg 
%the server \emph{uniformly at random samples} a subset of  $K$  clients (i.e., $K\! :=\! \left| \mathcal{K}^{r}\right|, \mathcal{K}^{r}\! \subseteq\!  \mathcal{N}$) and \emph{broadcast} the latest model $\mathbf{w}^r$ to the selected clients. 
%Second, each sampled client $i$ lets $\mathbf{w}_i^r=\mathbf{w}^r$ and runs $E$ steps\footnote{$E$ is originally defined as epochs of SGD in \cite{mcmahan2017communication}. In this paper, we denote $E$ as the number of local iteration for theoretically analysis.} of \emph{local stochastic gradient decent (SGD)} on~\eqref{lo_ob}, and compute a local update $\mathbf{w}_i^{r+1}$. % in parallel. %, {where $\mathcal{K}^{(r)} \subseteq  \mathcal{N}$}. 

%Last, the server \emph{weighted aggregated} their updated model parameters and compute a new global model $\mathbf{w}^{\tau+1}$ for the next round. %of these $\left|\mathcal{K}^{(r)}\right|$ clients are  by the server. 

The above process repeats for many rounds until the global loss converges. %{Let $R$ be the total number of rounds, then the total number of iterations for each device is $ER$.} 

Recent works have demonstrated the effectiveness of FedAvg with theoretical convergence guarantees  in various settings \cite{%li2018federated,  
li2019convergence,haddadpour2019convergence,karimireddy2019scaffold,yang2021achieving, qu2020federated}. {However, these works 
assume that the server samples  clients either uniformly at random or proportional to data size, %, such that the aggregated gradient updates, in expectation, are \emph{unbiased} stochastic versions of updates towards full participation. % and thus enjoys the same convergence properties as local-update SGD \cite{yu2018parallel,stich2018local,wang2018cooperative}.  
which  may slow down the  wall-clock time for convergence %with respect to wall-clock time %suffers from slow error convergence speed with respect to  wall-clock time
due to the straggling effect and non-i.i.d. data % within the uniformly sampled clients
\cite{stragglers}. Thus, a careful client sampling design should tackle both system and statistical heterogeneity for fast convergence.}
%Due to the characteristics of statistical and system heterogeneity,  {these sampling schemes can suffer from slow error convergence speed in terms of  wall-clock time \cite{stragglers}.}%, e.g., high convergence variance from non-i.i.d data, and .}  %due to the effect of the following statistical and systems heterogeneity.
%\begin{itemize}
 %   \item First, 
%\end{itemize}

\subsection{System Model of FL with Client Sampling $\mathbf{q}$} %$\mathbf{q}=\{q_1, \ldots, q_N\}$}
%\textbf{Need to Optimize Convergence in terms of Error versus Wall-clock Time.} 
%\subsubsection{Client Sampling Model} 
%In this work, %to tackle FL system and statistical heterogeneity, 
%we sample %$K$ %(i.e., $K := \left| \mathcal{K(\mathbf{q})}^{r}\right|$) out of $N$ clients 
We aim to sample clients according to a probability distribution 
$\mathbf{q}\!=\!\{q_i, \forall i\!\in\!\mathcal{N}\}$, where $0\!<\! q_i\!<\!1$   and $\sum_{i=1}^Nq_i\!=\!1$. Through  optimizing $\mathbf{q}$, we want to 
%study how to 
address system and statistical heterogeneity %via optimizing $\mathbf{q}$ %consider a novel client sampling strategy that can 
so as to minimize the wall-clock time for convergence. We describe the system model as follows. %We first state our key assumptions as follows. %  both system and statistical heterogeneity for speeding up the wall-clock training time while ensuring the convergence. %, denoted as $T_{tot}$.
%\subsubsection{System Model} 
%Unlike previous works of uniform sampling, %Similar to existing works \cite{li2018federated,li2019convergence,haddadpour2019convergence,yang2021achieving, karimireddy2019scaffold, qu2020federated}, we sample $K$ clients in each round $r$ (i.e., $K := \left| \mathcal{K}^{(r)} \right|$) where the sampling is uniform (without replacement) out of all $N$ clients.  
%We assume that in each round $r$, $K$  clients (i.e., $K := \left| \mathcal{K(\mathbf{q})}^{(r)}\right|$) are sampled from all $N$ clients according to sampling probability  $\mathbf{q}\!\!=\!\!\{q_1, \ldots, q_N\}$ $\left(\sum_{i=1}^Nq_i=1,  0\le q_i\le1,\  \forall i \in \{1,\ldots,N\}\right)$. %and  clients are agnostic to sampling strategy. 
\subsubsection{Sampling Model} 
Following recent works \cite{li2018federated,  
li2019convergence,haddadpour2019convergence,karimireddy2019scaffold,yang2021achieving, qu2020federated}, we assume %$K$ %($K := \left| \mathcal{K(\mathbf{q})}^{r}\right|$)
 that the server establishes the sampled client set $\mathcal{K(\mathbf{q})}^{r}$ by sampling $K$ times \emph{with replacement} from  the total $N$ clients, %according to probability $\mathbf{q}\!=\!\{q_1, \ldots, q_N\}$, %our wall-clock time and  convergence analysis in the next section.
where $\mathcal{K(\mathbf{q})}^{r}$ is a \emph{multiset} in which a client may appear more than once. The aggregation weight of each client $i$ is multiplied by the number of times it appears in $\mathcal{K(\mathbf{q})}^{r}$.
\subsubsection{Statistical Heterogeneity Model} We consider the standard FL setting where the training data are distributed in an unbalanced and  non-i.i.d. fashion among clients.   %we assume $F^*$ and $F_i^*$ for different client $i$ are different.
\subsubsection{System Heterogeneity Model}
%To characterize system heterogeneity, 
Following the same setup of \cite{wang2018adaptive} and \cite{stragglers}, {we denote $t_i$ as the round time of client $i$, which includes both local model computation time and global communication time.} %expected.
 %\subsubsection{FL Training Model} 
%\footnote{\textcolor{blue}{The round time $t_i$ may include both local model computation time and global communication time, whose impact will be analyzed in our future work.  %Future works may study the impact of dynamic communication time in each round, i.e., wireless communications with fading channels.
%We do not consider the time for model aggregation at the server.
%For clients with wireless communications, we could use the average channel gainof each client during the training time, which is commonly adopted in wireless FL \cite{tran2019federated,chen2020convergence,9207871}}. } 
{For simplicity, we assume that $t_i$ remains the same across different rounds for each client $i$, while for different clients $i$ and $j$, $t_i$ and $t_j$ can be different. The extension to time-varying $t_i$ is left for future work.} %To characterize the system heterogeneity, we assume $t_i$ is stable during the training process and follows a certain distribution, e.g.,  exponential distribution or uniform distribution.} 
Without loss of generality, as illustrated in Fig.~1, we sort all $N$ clients in the ascending order $\{t_i\}$, %in \eqref{Troundi} for all $N$ devices, %in a new sequence \emph{$\mathbf{{s}}$} 
% in an increasing fashion, %based on increasing of $t_k$, 
such that
\begin{equation}
\label{reorder}t_{1} \leq t_{2} \leq \ldots \leq t_{i}  \leq \ldots \leq t_{N}.\end{equation}  
\subsubsection{Total Wall-clock Time Model} 
We consider the mainstream synchronized FL model where each sampled client performs multiple (e.g., $E$) steps of local SGD before sending back their model updates to the server %for convergence analysis as 
(e.g., \cite{mcmahan2017communication, bonawitz2019towards,   
li2019convergence,haddadpour2019convergence,karimireddy2019scaffold,yang2021achieving, qu2020federated}).
For synchronous FL, the per-round time is limited by the slowest client (known as straggler).   %. Considering the heterogeneous communication delay , %heterogeneous computation and communication delay, 
%Similar to \cite{stragglers}, we denote $t_i, \forall i\in\{1,2,...,N\}$ as the%expected
%\footnote{We do not consider the time for model aggregation at the server, and assume the communication time is stable in each round. We will study the impact of dynamic communication time, i.e., wireless networks, in the future.} communication round time of client $i$ as illustrated in Fig.~1,  and without loss of generality, sort all $N$ clients in the ascending order $\{t_i\}$, %in \eqref{Troundi} for all $N$ devices, %in a new sequence \emph{$\mathbf{{s}}$} 
% in an increasing fashion, %based on increasing of $t_k$, 
 %for completing local computation and model communication. %Clearly, due to systems heterogeneity, clients with larger $t_i$ corresponds to slower clients or stragglers. 
%Because clients compute and communicate in parallel, for each round $r$, 
%the per-round time $T^{(r)}(\mathbf{q})$ depends on the sampled slowest client (also  straggler). %, which also depends on the sampling probability $\mathbf{q}=\{q_1, \ldots, q_N\}$. %\footnote{This is because in synchronized FL systems, the server needs to collect all updates from the sampled clients before performing global aggregation.}
Thus, the per-round time $T^{(r)}(\mathbf{q})$  of the entire FL process is %can be expressed as %By denoting the randomly sampled clients set as $\mathcal{K}^{(r)}$, where $\vert \mathcal{K}^{(r)} \vert =K$, we have 
\begin{equation}
\label{Tround}
T^{(r)}(\mathbf{q}):=\max _{i \in \mathcal{K}(\mathbf{q})^{(r)}}\left\{t_{i}\right\}.%, \ \forall k \in \{1, \ldots, N\}.
\end{equation}
%Therefore, the total learning time $T_\textnormal{tot}$ after $R$ rounds %$T$ iterations (equivalent to ${T}\slash E$ rounds) 
%is\footnote{We do not consider the time cost for model aggregation in Line~\ref{alg:fedavgStep4}, because it  needs much less time than that for local model updates and communications.}
%\begin{equation}
%\label{Ttot}
%T_\textnormal{tot}(\mathbf{q},R)=\sum\nolimits_{r=1}^{R}T^{(r)}(\mathbf{q})=\sum\nolimits_{r=1}^{R}\max _{i \in \mathcal{K}(\mathbf{q})^{(r)}}\left\{t_{i}\right\}.%, \ \forall k \in \{1, \ldots, N\}.
%\end{equation}
%The aim of FL is to guarantee that, after $R$ training rounds, the obtained final global loss $F\left(\mathbf{w}_R(\mathbf{q})\right)$  converges to the (true) minimum value $F^*$ with desired error precision $\epsilon$, where  $R(\mathbf{q})$ and the global model $\mathbf{w}(\mathbf{q})$ both depend on the sampling probability $\mathbf{q}$. 
%global model $\mathbf{w}(\mathbf{q},R)$ is function of sampling probability $\mathbf{q}$ because the sampled clients could have different data quality due to non-i.i.d. data, which depends how clients are sampled in each round.  
Therefore, the total learning time $T_\textnormal{tot}(\mathbf{q},R)$ after $R$ rounds %$T$ iterations (equivalent to ${T}\slash E$ rounds) 
is %can be expressed as%\footnote{}
\begin{equation}
\label{Ttot}
T_\textnormal{tot}(\mathbf{q},R)=\sum\nolimits_{r=1}^{R}T^{(r)}(\mathbf{q})=\sum\nolimits_{r=1}^{R}\max _{i \in \mathcal{K}(\mathbf{q})^{r}}\left\{t_{i}\right\}.%, \ \forall k \in \{1, \ldots, N\}.
\end{equation}

\subsection{Problem Formulation} 
Our goal is to %obtain optimal sampling probability $\mathbf{q}=\{q_1, \ldots, q_i, \ldots, q_N\}$ that
minimize the {expected} total learning time $\Expect[T_\textnormal{tot}(\mathbf{q},R)]$, while ensuring that the expected global loss  $\Expect[F\left(\mathbf{w}^R(\mathbf{q})\right)]$ converges to the minimum value $F^{*}$ with an $\epsilon$ precision, 
with $\mathbf{w}^R(\mathbf{q})$ being the aggregated global model after $R$ rounds with client sampling probabilities $\mathbf{q}$. % under the proposed adaptive sampling $\mathbf{q}$, % performance of any desired precision $\epsilon$, 
This translates into the following problem:
\begin{equation}\begin{array}{cl}
\label{ob1}
\!\!\!\!\!\!\!\textbf{P1:}\quad \min_{\mathbf{q}, R} & \Expect[T_\textnormal{tot}(\mathbf{q},R)] \\
\quad\quad \text { s.t. } & \Expect[F\left(\mathbf{w}^R(\mathbf{q})\right)]-F^{*} \le \epsilon,\\
&\sum_{i=1}^Nq_i=1, \\
& q_i>0,  \forall i \in \mathcal{N}, %\{1,\ldots N\},
\ R \in \mathbb{Z}^{+}. 
\end{array}\end{equation}
%where $\Expect[F\left(\mathbf{w}_R(\mathbf{q})\right)]$ is the expected loss after $R$ rounds, $F^*$ is the true minimum value of $F$, and $\epsilon$ is the desired error precision. 
 The expectation in $\Expect[T_\textnormal{tot}(\mathbf{q},R)]$ and $\Expect[F\left(\mathbf{w}^R(\mathbf{q})\right)]$ in \eqref{ob1} is due to the randomness in client sampling $\mathbf{q}$ and local SGD. % in each round. %Source of randomness: In our analysis, there are two sources of randomness. One results from the SGD and the other is from the random sampling of devices. We take the expectation to erase those randomness.
Solving Problem \textbf{P1}, however, is challenging in two aspects:
\begin{enumerate}
    \item It is generally impossible to find out how  $\mathbf{q}$ and  $R$  affect the final model $\mathbf{w}^R(\mathbf{q})$ and the corresponding loss function  $\Expect[F\left(\mathbf{w}^R(\mathbf{q})\right)]$ before actually training the model. % conducting the training process.
    Hence, we need to obtain an analytical expression with respect to $\mathbf{q}$ and  $R$  to predict how they affect $\mathbf{w}^R(\mathbf{q})$ and $\Expect[F\left(\mathbf{w}^R(\mathbf{q})\right)]$. %, e.g., via a convergence bound, 
   % to relate $\mathbf{q}$ and  $R$ with $\Expect[F\left(\mathbf{w}^R(\mathbf{q})\right)]$.
    %It is difficult to obtain an analytical expression to represent the convergence constraint in  \textbf{P1} for arbitrary client sampling probability $\mathbf{q}$ and training round $R$. 
    
    %convergence bound that can adapt to arbitrary client sampling probability $\mathbf{q}$ and training round $R$ with the convergence constraint.

\item %The objective function  $\Expect[T_\textnormal{tot}(\mathbf{q},R)]$  is difficult to optimize due to the \emph{non-linear maximum} function (straggling effect) in \eqref{Ttot}

The objective  $\Expect[T_\textnormal{tot}(\mathbf{q},R)]$ is complicated to optimize due to the straggling effect in \eqref{Ttot}, %related to the set of sampled clients %with adaptive sampling
%in \eqref{Ttot}, 
which can result in %includes both polynomial and exponential terms of %is a polynomial function of 
%$q_1,\ldots,q_N$ and  %with $\mathcal{O}(N^K)$ terms, 
%which is complicated to solve because $N$ is usually very large in cross-device FL settings.
%Moreover, we show that Problem \textbf{P1} 
%can be
a non-convex optimization problem even for simplest cases as we will show later.%, e.g., $K=1$.%  in \eqref{Ttot}, whose with the order of $N^K$ terms 

%objective function $\Expect[T_\textnormal{tot}(\mathbf{q},R)]$  is difficult to optimize due to the \emph{non-linear maximum} function (straggling effect) in \eqref{Ttot}, whose analytical form is a polynomial function of $\mathbf{q}$ with the order of  $N^K$ terms, and we later show that the formulated problem can be non-convex to solve even for some simple cases. 

%formulate an analytical expression of  $\Expect[T_\textnormal{tot}(\mathbf{q},R)]$ for the coupled sampling probability $\mathbf{q}$ and training round $R$, especially due to the \emph{non-linear maximum} function in the round time $T^{(r)}$.  

%establish an analytical relationship between the expected total learning time $\Expect[T_\textnormal{tot}]$ and the %convergence constraint with 
%$\mathbf{q}$ and $R$ in the \emph{coupled} convergence constraint,  especially due to the \emph{non-linear maximum} function in the round time $T^{(r)}$.
\end{enumerate} 
%it is difficult to establish an analytical expression of total learning time with respect to sampling probability.
%find an \emph{exact analytical expression} to relate  sampling probability $\mathbf{q}$ and training round $R$
%with total learning time $T_\textnormal{tot}$, especially due to the \emph{non-linear maximum} function in the round time $T^{(r)}$. %\eqref{Ttot}. 
%Second, it is difficult to establish an find an tractable convergence result that analytical connects $\mathbf{q}$ and $R$  with the convergence constraint.} 

In Section~\ref{sec:convergence} and Section~\ref{sec:optimizationProblem}, we address these two challenges, respectively, and propose approximate  algorithms to find an approximate solution to % efficiently solve 
Problem \textbf{P1} efficiently.   %In the following section, we propose an algorithm that approximately solves P1, which we later show with extensive experiments that the proposed solution can achieve a near-optimal performance of P1.} %$\Expect[F(\mathbf{w}_T)]$ and $\epsilon$. 
%In the following section, we propose an algorithm that approximately solves \textbf{P1}, which we later show with extensive experiments that the proposed solution can achieve a  near-optimal performance of \textbf{P1}.}
%by analyzing the sampling properties we successfully established a combinatorial expression for $t_\textnormal{tot}$ in terms of $E$, $K$ and $T$.  An efficient control algorithm is developed to achieve the approximate solution, which, nevertheless, we later show has a near-optimal performance.
%$C_\textnormal{tot}$ and the control variables
% \footnote{We use this convergence bound as it characterize the generic problem settings in FL community}

\section{Convergence Bound for Arbitrary Sampling}
\label{sec:convergence}
In this section, we address the first challenge by % shows how to %approximately
deriving a new tractable convergence bound  for  arbitrary 
%establishes the connection between the convergence constraint with 
client sampling probabilities. % adapt to differnt K and E, 
%This  section  shows  how  to  obtain  the  error-convergencebound  that  adapts  to  the  arbitrary  sampling  probability,  toaddress the first challenge

%\noindent
%\textbf{Remark:} Based on the above results in Theorem 1 and 2, %we have decoupled and analytically obtained the expression for $\Expect[T_\textnormal{tot}(\mathbf{q},R)]$ with $\mathbf{q}$ and $R$. 
%the objective function $\Expect[T_\textnormal{tot}]$ in \textbf{P1}
%can be analytically expressed as
%\begin{equation}
%\label{obj_fun_new} \Expect[{T_\textnormal{tot}}]\!= \!  \sum_{i=1}^{N}\!\left[\left(\sum\nolimits_{j=1}^{i} q_{j}\right)^{K}\!\!-\!\left(\sum\nolimits_{j=1}^{i-1} q_{j}\right)^{K}\right] \cdot t_{i}R(q).% \frac{T}{E}.
%\end{equation}

%\subsection{New Convergence Bound with Adaptive Sampling $\mathbf{q}$} % 
%This subsection shows how to establish the relationship between the convergence constraint with sampling probability $\mathbf{q}$.
\subsection{{Machine Learning Model Assumptions}} To ensure a tractable convergence analysis, %as adopted in \cite{li2019convergence, qu2020federated}
we first state several assumptions on the local objective functions ${F_i}\left( \mathbf{w} \right)$.
\begin{assumption}
L-smooth: For each client $i\! \in\! \mathcal{N}$, $F_{i}$ is  $L$-smooth, i.e.,  
$\|\nabla\! f(\mathbf{v})\!-\!\nabla\! f(\mathbf{w})\| \!\leq\! L\|\mathbf{v}\!-\!\mathbf{w}\|$  for all $\mathbf{v}$ and $\mathbf{w}$.
%F_{i}(\mathbf{v}) \leq F_{i}(\mathbf{w})+(\mathrm{v}-$
%$\mathbf{w})^{T} \nabla F_{i}(\mathbf{w})+\frac{L}{2}\|\mathbf{v}-\mathbf{w}\|_{2}^{2}$.
\end{assumption}
\begin{assumption}
Strongly-convex:  For each client $i \in \mathcal{N}$, $F_{i}$ is $\mu$-strongly convex, i.e., $F_{i}(\mathbf{v}) \geq F_{i}(\mathbf{w})+(\mathbf{v}-$
$\mathbf{w})^{T} \nabla F_{i}(\mathbf{w})+\frac{\mu}{2}\|\mathbf{v}-\mathbf{w}\|_{2}^{2}$ for all $\mathbf{v}$ and $\mathbf{w}$.
\end{assumption}
\begin{assumption}
Bounded local variance: %Let $\xi^{(i)}$ be local data sampled selected from the $i$-th client uniformly at random.
For each device $i\in \mathcal{N}$,  %at any iteration $\tau$,
the variance of its  stochastic gradient is bounded: $\mathbb{E}\left\|\nabla F_{i}\left(\mathbf{w}_i, \xi_{i}\right)\!-\!\nabla F_{i}\left(\mathbf{w}_i\right)\right\|^{2} \leq \sigma_{i}^{2}$.
 %Let $\sigma^{2}:=\sum_{i=1}^{N} p_{i} \sigma_{i}^{2}.$
\end{assumption}
\begin{assumption}
Bounded local gradient: For each client $i \in \mathcal{N}$, the expected squared norm of stochastic gradients is bounded: $\mathbb{E}\left\|\nabla F_{i}\left(\mathbf{w}_{i}, \xi_{i}\right)\right\|^{2} \leq G_i^{2}$.
\end{assumption}
Assumptions 1--3 are %satisfied by a range of popular objective functions such as $\ell_{2}$-norm regularized linear regression, logistic regression, and softmax classifier.
%Nevertheless, the experimentation results that will be presented in Section VI show that our results also works well for non-convex function such as neural network. %Assumptions 3 and 4 have been 
common in many existing studies of  convex FL problems, such as  $\ell_{2}$-norm regularized linear regression, logistic regression (e.g., \cite{li2019convergence,yu2018parallel,chen2020optimal,stich2018local,qu2020federated,cho2020client}).  %Assumption 3 can be removed if clients perform full gradient when their local datasize is not large. 
Nevertheless, the experimental results to be presented
in Section VI show that our approach also works well for \emph{non-convex}
loss functions. 
Assumption 4, however, is a less restricted version of the assumption made in \cite{li2019convergence,yu2018parallel,chen2020optimal,stich2018local,qu2020federated,cho2020client}, where those studies have assumed that $G_i$ is uniformly bounded by a universal $G$. Instead, we allow each client $i$ to have a unique $G_i$, which yields our optimal client sampling design as we will show later. 

%We will show later that this definition allows for a tighter convergence bound under %, and also enlightens us to distinguish clients' data quality for optimal
%client sampling $\mathbf{q}$.% design. % in the following subsections. %heterogeneity. This, we show in the following subsections, leads to a tighter convergence bound.% by characterizing data heterogeneity.

%how $G_i$ and the corresponding $q_i$ affect the FL convergence rate. %can be used to reflect clients' heterogeneous data distribution and guide our optimal sampling strategy.  

\subsection{{Aggregation with Arbitrary Client Sampling Probabilities}} 
This section shows how to aggregate clients' model updates under sampling probabilities $\mathbf{q}$, such that the aggregated global model is unbiased compared to that with full client participation, which leads to our  convergence result. 

We first define the \emph{virtual weighted aggregated model with full client participation} in round $r$ as
\begin{equation}
    \label{full_sample}
    \overline{\mathbf{w}}^{r+1}:=\sum\nolimits_{i=1}^Np_i\mathbf{w}_i^{r+1}.
\end{equation}
With this, we can derive the following result. %\newtheorem{lemma}{{Lemma}}
\begin{lemma}
\label{adaptive_sam_agg}
\textbf{(Adaptive Client Sampling and Model Aggregation)} %Let $\overline{\mathbf{w}}_{r+1}=\sum_{i=1}^Np_iw_{r+1}^{(i)}$ be the virtual aggregated global model update with full participation based on weight $p_i$, then, 
When clients $\mathcal{K}(\mathbf{q})^{r}$ are sampled with probability $\mathbf{q}=\{q_1, \ldots q_N\}$ and their local updates are aggregated as %\begin{equation}
 %   \label{aggregation}
 $\mathbf{w}^{r+1} \leftarrow \mathbf{w}^{r}+\sum_{i \in \mathcal{K}(\mathbf{q})^{r}} \frac{p_{i}}{K q_{i}} \left(\mathbf{w}_i^{r+1}-\mathbf{w}^{r}\right)$,
%\end{equation}
we have  
\begin{equation}
    \label{unbiased_agg}
    \Expect_{\mathcal{K}(\mathbf{q})^{r}}[\mathbf{w}^{r+1}]= \overline{\mathbf{w}}^{r+1}.
\end{equation}
\end{lemma}
\noindent\emph{Proof  Sketch.}
 The basic idea is to take expectation over the aggregated global model of the sampled clients $\mathcal{K}(\mathbf{q})^{r}$, and with some mathematical derivations, we have \eqref{unbiased_agg}. \qquad \qquad \qedsymbol

%\cite{zhao2015stochastic,needell2014stochastic}.
\textbf{Remark}: {The key insight %interpretation
of our {sampling} and {aggregation} is that {since we sample different clients with different probabilities (e.g., $q_i$ for client $i$),  \emph{we need to inversely re-weight  their updated model in the aggregation step (e.g.,  $\frac{1}{q_i}$ for client $i$), such that the aggregated  model is still unbiased towards that with full client participation}.}} 
We summarize how  the  server  performs  client  sampling  and model  aggregation in 
Algorithm~\ref{FL_sample_agg}, where the main differences compared to the de facto FedAvg  in \cite{mcmahan2017communication} are the \emph{Sampling} (Line~\ref{alg:adaptivefedavgStep1}) and \emph{Aggregation} (Line~\ref{alg:adaptivefedavgStep4}) procedures. Notably, Algorithm~\ref{alg:adaptivefedavg} recovers FedAvg algorithm with uniform  sampling  when letting  $q_i=\frac{1}{N}$, or with weighted sampling when letting $q_i=p_i$ in \cite{li2019convergence}.

% Algorithm 1 recovers the FedAvg algorithm .%previous sampling schemes 
%in the sense that when $q_i=\frac{1}{N}$ or $q_i=p_i$, our Algorithm 1 recovers the sampling schemes in . %both of which have been proved with convergence guarantee \cite{li2019convergence}. %,that adapts to any sampling probability $\mathbf{q}$.  %This idea is also denoted as \emph{importance sampling}, which is adopted for improving the convergence rate of SGD \cite{zhao2015stochastic,needell2014stochastic}.  

        %: The server randomly samples   $\mathcal{K}^{(r)}$ client according to   distribution $\mathbf{q}$; (corresponds to Lines~\ref{alg:adaptivefedavgStep1})
    %: The server computes the new global model parameter with $\mathbf{w}_{t+E} \leftarrow \mathbf{w}_{t}-\sum_{k \in \mathcal{S}_{t}} \frac{p_{k}}{K q_{k}} \left(\mathbf{w}_{t+E}^{k}-\mathbf{w}_{t}\right)$;  (corresponds to Lines~\ref{alg:adaptivefedavgStep4})

\begin{algorithm}[t]
\label{FL_sample_agg}
\small
	\caption{FL with Arbitrary Client Sampling}
	\label{alg:adaptivefedavg}
	\KwIn{Sampling probabilities $\mathbf{q}=\{q_1, \ldots, q_N\}$, $K$, $E$, precision $\epsilon$, initial model $\mathbf{w_0}$}
	\KwOut{Final model parameter $\mathbf{w}^R$}
	\For{$r\leftarrow0,1,2,..., R$%\frac{T}{E} $
	}{	\emph{Server randomly samples a subset of clients  $\mathcal{K}(\mathbf{q})^{r}$ {according to}   $\mathbf{q}$, and sends current global model   $\mathbf{w}^r$ to the selected clients\label{alg:adaptivefedavgStep1}\tcp*{\textbf{Sampling}}}

Each sampled client $i$ %\in \mathcal{K}(\mathbf{q})^{r} $ %updates $\mathbf{w}_r$ by running 
lets $\mathbf{w}_i^{r,0}\!\leftarrow\!\mathbf{w}^r$, and  performs %$E$ steps local SGD on \eqref{lo_ob}:
%\For{$j=0,1,2,..., E-1$}{
$\mathbf{w}_i^{r,j+1} \!\leftarrow\! \mathbf{w}_i^{r,j}-\eta^{r} \nabla F_{k}\left(\mathbf{w}_i^{r,j}, \xi_i^{r,j}\right),j\!=\!0,1,\dots, E\!-\!1$,
and lets $\mathbf{w}_i^{r+1}\leftarrow\mathbf{w}_i^{r,E}$		\label{alg:adaptivefedavgStep2}\tcp*{Computation}
		
	Each sampled client $i$ % \in \mathcal{K}(\mathbf{q})^{r}$  
	sends back updated model $\mathbf{w}_i^{r+1}$ to the server
	\label{alg:adaptivefedavgStep3}\tcp*{Communication}

\emph{Server computes a new global model parameter as  $\mathbf{w}^{r+1} \leftarrow \mathbf{w}^{r}+\sum_{i \in \mathcal{K}(\mathbf{q})^{r}} \frac{p_{i}}{K q_{i}} \left(\mathbf{w}_i^{r+1}-\mathbf{w}^r\right)$ \label{alg:adaptivefedavgStep4}\tcp*{\textbf{Aggregation}}}

%		$r \leftarrow r+1$;
	}
\end{algorithm}
\subsection{{Main Convergence Result for Arbitrary Client Sampling}} 
Based on %the proposed adaptive sampling and aggregation scheme in
%Algorithm~\ref{alg:adaptivefedavg} and
Lemma~\ref{adaptive_sam_agg}, we present the main convergence result for arbitrary client sampling in Theorem~\ref{convergencebound}.

\begin{theorem}\label{convergencebound}
\textbf{(Convergence Upper Bound)} Let Assumptions 1 to 4 hold, % given  client sampling probability $\mathbf{q}=\{q_1, \ldots, q_N\}$ and model aggregation described in Lemma~\ref{adaptive_sam_agg}, %Let Assumptions 1 to 4 hold and $L, \mu, \sigma_{i}, G_i$ be defined therein. 
$\gamma=\max \{\frac{8 L}{\mu}, E\}$, and decaying learning rate $\eta_{r}=\frac{2}{\mu(\gamma+r)}$. %($\tau$ is the iteration index), 
For given client sampling probabilities $\mathbf{q}=\{q_1, \ldots, q_N\}$ and the corresponding aggregation described in Lemma~\ref{adaptive_sam_agg}, the optimality gap after $R$ rounds satisfies % with $K$ sampling clients  participation and $E$ local iterations we have
\begin{equation}
    \label{convergence}
    \begin{array}{c}
     \Expect[F\left(\mathbf{w}^R(\mathbf{q})\right)]-F^{*} \le\frac{1}{R}\left(\alpha{\sum_{i=1}^N\frac{p_i^2G_i^{2}}{q_i}} +\beta\right),
    \end{array}
\end{equation}
where $\alpha\!=\!\frac{8LE}{\mu^2K}$ and $\beta\!=\!\frac{2L}{\mu^2E}B+\frac{12L^2}{\mu^2E}\Gamma+ \frac{4L^2}{\mu E}\left\|\mathbf{w}_{0}\!-\!\mathbf{w}^{*}\right\|^{2}$, with $B\!=\!\sum\limits_{i=1}^{N} p_{i}^{2} \sigma_{i}^{2}\!+\!8\!\sum\limits_{i=1}^Np_iG_i^2E^{2}\!$ and $\Gamma\!=\!F^{*}\!-\!\sum_{i=1}^{N} p_{i} F_{i}^{*}$.  %are independent to sampling probability $\mathbf{q}$.
\end{theorem}
\noindent\emph{Proof  Sketch.}
%Our convergence bound with adaptive sampling 
First, following the similar proof of convergence under full client participation in \cite{stich2018local,li2019convergence}, we show that

%\begin{equation}
 %   \label{convergence_full}
 $\Expect[F\left(\overline{\mathbf{w}}^R)\right)]-F^{*} \le\frac{\beta}{R}$,
%\end{equation}
where $\Expect[F\left(\overline{\mathbf{w}}^R)\right)]$ is the  expected global loss after $R$ rounds with full participation, and $\beta$ is the same as in \eqref{convergence}.  %$\beta^{\prime}\!=\!\beta\!+\!\frac{16L(E\!-\!1)^{2}}{\mu^2E}\left(\sum_{i=1}^Np_iG_i^2\!-G^2\!\right)$ with $G=\max G_i$ being the  uniform bound of stochastic gradient norm for all clients. % as in Assumption~4. 
%Note that the expectation in \eqref{convergence_full} is taken over from stochastic gradients without client sampling.
Then, for client sampling probabilities  $\mathbf{q}$, as we have shown that the expected aggregated global model $ \Expect_{\mathcal{K}(\mathbf{q})^{r}}[\mathbf{w}^{r+1}]$ is unbiased compared to full participation $\overline{\mathbf{w}}^{r+1}$ in Lemma~\ref{adaptive_sam_agg}, we can show that the expected difference  of the two (sampling variance) is bounded as follows: 
\begin{equation}
    \label{bounded_variance}
        \begin{array}{c}
            \mathbb{E}_{\mathcal{K}(\mathbf{q})^{r}}\left\|\mathbf{w}^{r+1}-\overline{\mathbf{w}}^{r+1}\right\|^{2} \leq \frac{4}{K}  \sum_{i=1}^N\frac{p_i^2G_i^{2}}{q_i}\left({\eta^{r}} E\right)^{2}.
        \end{array}
\end{equation}
After that we use induction to obtain a non-recursive bound on $\mathbb{E}_{\mathcal{K}(\mathbf{q})^{r}}\left\|\mathbf{w}^R-{\mathbf{w}}^*\right\|^{2}$,
which is converted to a bound on  $\Expect[F\left(\mathbf{w}^R(\mathbf{q})\right)]-F^{*}$ using $L$-smoothness. Finally, we show that the main difference of the contraction bound compared to full client participation is the sampling variance in \eqref{bounded_variance}, which yields the additional term of  $\alpha{\sum_{i=1}^N\frac{p_i^2G_i^{2}}{q_i}}$ in \eqref{convergence}.   \qquad \qquad \qedsymbol
%, with the result in  Lemma~\ref{adaptive_sam_agg} that , the main difference between the convergence bound with full participation in \eqref{convergence_full} is the % only need to show that the
%\begin{equation}
 %   \label{unbiased_agg_R}
  %  \Expect_{\mathcal{K}(\mathbf{q})^{r}}[\mathbf{w}^{r}]= \overline{\mathbf{w}}^{r}.
%\end{equation}
%although our Lemma~\ref{adaptive_sam_agg} guarantees the expected aggregated model update $\mathbf{w}^{r+1}(\mathbf{q})$ is unbiased towards the model with full participation $\overline{\mathbf{w}}^{r+1}$, 
%the client sampling procedure will introduce a 
%resulting sampling variance  %$\mathbb{E}_{\mathcal{K}(\mathbf{q})^{r}}\left\|\mathbf{w}^{r+1}(\mathbf{q})-\overline{\mathbf{w}}^{r+1}\right\|^{2}$,  
%is bounded by $\alpha{\sum_{i=1}^N\frac{p_i^2G_i^{2}}{q_i}}$, which concludes this proof. % term in our convergence bound in \eqref{convergence}.
%We give the full proof in Appendix A.\footnote{The proof is inspired by the results for uniform sampling in \cite{li2019convergence}. }  are independent to sampling probability $\mathbf{q}$

%\textbf{Remark:} %The main observation of the derived convergence upper bound can be summarized as follows.
%\begin{itemize}
 %   \item 
 Our convergence bound in \eqref{convergence} characterizes the relationship between client sampling probabilities $\mathbf{q}$ and the number of rounds $R$ for reaching the target precision ($\Expect[F\left(\overline{\mathbf{w}}^R)\right)]-F^{*} \le\epsilon$). Notably, our bound %can be seen as a generalization result for arbitrary client sampling, as our bound 
 generalizes  the convergence results in \cite{li2019convergence}, where clients are uniformly sampled ($q_i\!=\!\frac{1}{N}$) or weighted sampled ($q_i\!=\!p_i$). %, our convergence bound recovers the convergence results in .
      %\item Our convergence bound adapts to standard FL with different configuration parameters $E$ and $K$, which can be jointly optimized for further reducing total learning time, since a larger $E$ may increase local computation time, and a larger $K$ can affect communication time in wireless networks \cite{luo2020cost}.  
  %  \item 
Moreover, our convergence bound also motivates the optimal client sampling design for homogeneous systems, i.e., all clients with the same communication time $t_0$, as follows. 
%explains why uniformly sampling or weighted sampling can yield slow convergence, as $q_i=\frac{1}{N}$ or $q_i=p_i$ cannot minimize the convergecne upper bound  not well match the corresponding coefficients $p_i^2G_i^2$.  which motivates the following optimal sampling design
%for improving error convergence with respect to the number of rounds, 
%which we show with Corollary~\ref{opt_q_statis}}.
 %  \end{itemize}
 %\newtheorem{corollary}{Corollary}
\begin{corollary}
\label{opt_q_statis}
%\textbf{(Optimal Sampling for Speeding up FL with respect to the Number of Rounds)} 
{For FL with homogeneous communication time i.e., $t_i=t_0$, for all  $i \in \mathcal{N}$, the optimal client sampling probabilities $\mathbf{q}$ for Problem \textbf{P1} is} %P1minimizing $\Expect[T_\textnormal{tot}(\mathbf{q},R)]$  %the number of rounds $R$ 
% for reaching a given precision $\epsilon$ is given by %For a given  precision $\epsilon$ 
\begin{equation}
    \label{optsampling_statis}
        \begin{array}{c}
            q_{i}^{*}=p_{i} G_{i} \left/ \sum_{j=1}^{N} p_{j} G_{j}\right. .
        \end{array}
\end{equation}
\end{corollary}
\begin{prf}
{When clients have the same communication time, the round  time $T^{(r)}(\mathbf{q})$ in \eqref{Tround} is fixed as $t_0$, and thus \emph{minimizing the total learning time $\Expect[T_\textnormal{tot}(\mathbf{q},R)]$ in Problem \textbf{P1} is equivalent to minimizing the total number of  communication rounds} for reaching the target precision $\epsilon$.} Hence, by letting $\Expect[F\left(\mathbf{w}(\mathbf{q},R)\right)]-F^{*}=\epsilon$, and by moving $R$ in \eqref{convergence} from the right hand side to the left hand side of inequality, we have 
\begin{equation}
\label{corollary1_2}
    \begin{array}{c}
      R \le\frac{1}{   \epsilon}\left(\alpha {\sum\limits_{i=1}^N\frac{p_i^2G_i^{2}}{q_i}} +\beta\right).
    \end{array}
\end{equation} Thus, for a target precision $\epsilon$, computing  the optimal sampling $\mathbf{q}$ for minimizing the upper bound of $R$ is equivalent to solving 
\begin{equation}
    \label{sub_prob}
    \begin{array}{cl}
\quad \min_{\mathbf{q}} & \quad \  \sum_{i=1}^N\dfrac{p_i^2G_i^{2}}{q_i} \\
\quad \text { s.t. } &\sum_{i=1}^Nq_i=1, \ q_i>0,  \forall i \in \mathcal{N}.
\end{array}
\end{equation}
The problem in \eqref{sub_prob} can be easily solved with the Lagrange multiplier method in closed form as shown in  \eqref{optsampling_statis}. 
\end{prf}
%\textbf{Remark:} %Corollary~\ref{opt_q_statis} shows that, for minimizing the total number of communication rounds, %speeding up FL with respect to the number of rounds, the optimal client sampling probability  %without considering heterogeneous physical time, the optimal sampling probability for client $i$ 
%is proportional to the product of clients' data size ($p_i$) and local SGD norm  ($G_i$),\footnote{ Corollary~\ref{opt_q_statis} can be seen as an offline result of \cite{chen2020optimal}, whose optimal sampling is weighted by the norm of local stochastic gradient in each round, thus frequently requiring the knowledge of stochastic gradient from all clients.}     
%when $G_i$ is replaced with the online SGD norm in each round, our sampling result reduces to the optimal online sampling in \cite{chen2020optimal}.   
%which characterizes the impact and interplay between \emph{data quantity} and \emph{data quality}. % on client sampling. % affects the convergence speed with different sampling schemes, 
%For instance, sampling clients with large data size may not be optimal as they might have poor data quality (e.g., with only one class in a multi-class classification task). 

%\textbf{Remark:}{The client sampling strategy in Corollary~\ref{opt_q_statis} is the optimal solution of Problem \textbf{P1} only for the special case of homogeneous communication time. 
In the following, we show how to leverage the derived convergence bound in Theorem~\ref{convergencebound} to design the optimal client sampling for the general heterogeneous system of Problem \textbf{P1}. %with straggling effect. %only speeds up the error convergence with respect to the number of training rounds via adapting to statistical property. 
%\emph{To speed up the true error convergence rate with respect to wall-clock time}, we need to consider the physical round time, and further optimize client sampling  via jointly leveraging the property of system and statistical heterogeneity.  %instead of the number of iterations \emph{can be suboptimal with respect to minimizing wall-clock time due to the lack of system heterogeneity consideration}. %In the following section, we show how to utilize the convergence bound in Theorem~\ref{convergencebound} to achieve the optimal sampling probability $q$ that tackles both system and statistical heterogeneity.  

 \section{Optimal Adaptive Client Sampling Algorithm}
\label{sec:optimizationProblem}
In this section, we first %address the second challenge by 
obtain the analytical expression of  the expected total learning time $\Expect[T_\textnormal{tot}]$ with sampling probabilities $\mathbf{q}$ and training round $R$. %establishing the analytical relationship between the expected total learning time $\Expect[T_\textnormal{tot}]$ and the \emph{coupled} sampling probability $\mathbf{q}$ and training round $R$ in the convergence bound in %\eqref{convergence}.
%Theorem~\ref{convergencebound}. 
Then, %we formulate the analytical relationship between the expected total time $\Expect[T_\textnormal{tot}]$ and $\textbf{q}$ and $R$ with the convergence result. % using the knowledge of \emph{order statistics} 
we formulate an %alternative problem that includes an 
  approximate problem of the original Problem \textbf{P1} based on the convergence upper bound in %\eqref{convergence}.
Theorem~\ref{convergencebound}.  %which we show is non-convex but can be efficiently solved with global optimum guarantees. Finally, 
Finally, we develop an efficient algorithm to solve the new problem with insightful sampling principles. %learn the convergence-related unknown parameters, which leads to 
%, which achieves faster convergence time compared to unbiased uniform sampling in our .
%with marginal estimation overhead. %which can be efficiently solved after estimating % by obtaining 
%unknown parameters associated with the convergence bound. Finally, we propose online and offline algorithms to learn these unknown parameters, which can be adopted for different FL system settings.

\subsection{Analytical Expression for $\Expect[T_\textnormal{tot}(\mathbf{q})]$}
%\subsubsection{\textbf{Decouple $\Expect[T_\textnormal{tot}(\mathbf{q})]$}}

%We first analytically establish the expected total learning time $\Expect[T_\textnormal{tot}(\mathbf{q},R)]$ with the sampling probability $\mathbf{q}$.  %{Without loss of generality, we sort all $N$ clients in the descending order $\{t_i: \forall i\in\{1,2,...,N\}\}$, %in \eqref{Troundi} for all $N$ devices, %in a new sequence \emph{$\mathbf{{s}}$} 
% in an increasing fashion, %based on increasing of $t_k$, 
%such that}
%\begin{equation}
%\label{reorder}t_{1} \leq t_{2} \leq \ldots \leq t_{i}  \leq \ldots \leq t_{N}.\end{equation}
%
\begin{theorem}
\label{lemma:expect_tr}
%With the reordered $t_i$ as in \eqref{reorder}, 
The expected total learning time $\Expect[T_\textnormal{tot}(\mathbf{q},R)]$ is %$\Expect[T^{(r)}(\mathbf{q})]$ 
%can be expressed as% expectation of $\Expect[T^{(r)}(\mathbf{q})]$ in \eqref{etot} can be expressed as
\begin{equation}
%\begin{array}{cl}
\label{E_T_total}
%\Expect[T^{(r)}(\mathbf{q})]
    \begin{array}{c}
        \!\!\!\Expect[T_\textnormal{tot}(\mathbf{q},\!R)]\!=\!\sum_{i=1}^{N}\!\left[\left(\sum_{j=1}^{ i} q_{j}\right)^{\!K}\!\!\!-\!\left(\sum_{j=1}^{i-1} q_{j}\right)^{\!K}\right]  t_{i}  R.\!\!\!
   %&=e_{p}\frac{KT}{N}+2e_{m}\frac{KT}{NE}
   \end{array}\!\!\!
\end{equation}
\end{theorem}%\end{lemma}
\noindent\emph{Proof  Sketch.}
The idea is to show that with sampling probabilities $\mathbf{q}$, the expected per-round time $\Expect[T^{(r)}(\mathbf{q})]$  in \eqref{Tround} is
\begin{equation}
\label{Eperround_T}
    \begin{array}{c}
        \Expect[T^{(r)}(\mathbf{q})]=\sum_{i=1}^{N}\!\left[\left(\sum\nolimits_{j=1}^{ i}\! q_{j}\right)^{\!K}\!\!-\!\left(\sum\nolimits_{j=1}^{i-1} \!q_{j}\right)^{\!K}\right]  t_{i}.
    \end{array}
\end{equation}
We first show that  the probability of client $i$ being the slowest one (e.g., straggler) amongst the $K$ sampled clients in each round is $(\sum\nolimits_{j=1}^{i} q_{j})^{K}\!-\!(\sum\nolimits_{j=1}^{i-1} q_{j})^{K}$. %The first term  $(\sum\nolimits_{j=1}^{i} q_{j})^{K}$ represents the probability of the sampled $K$ clients all from the reordered clients sequence $\{1,2,\ldots,i\}$, while  $(\sum\nolimits_{j=1}^{i-1} q_{j})^{K}$ is the  probability of the sampled $K$ clients all from reordered clients sequence $\{1,2,\ldots,i-1\}$. Thus, the difference of the two terms is the probability that amongst the sampled $K$ clients, at least one client is client $i$, which is the slowest in the sampled set. 
Since we sample devices according to $\mathbf{q}$,
taking the expectation of all $N$ clients over time $q_i$ gives % all the  gives Summing up the weighted time $t_i$  over all $N$ clients leads to
\eqref{Eperround_T}, and for $R$ rounds we have  \eqref{E_T_total}. \qquad \quad \  \qedsymbol
%Then, given that the round number $R$ is only determined in the training procedure, which is independent to the wall-clock time in each round,  for $R$ training rounds, we have \eqref{E_T_total}. %the expected total learning time is \eqref{E_T_total}, which concludes the proof.    %is indenpendent to the round time Then, we show the expected per-round time 

\subsection{Approximate Optimization Problem for Problem \textbf{P1}}
Based on Theorem~\ref{lemma:expect_tr}, %the above analytical expression of the expected total learning time $\Expect[T_\textnormal{tot}]$ in \eqref{E_T_total} and %the analytical convergence upper bound in \eqref{convergence} in Theorem~\ref{convergencebound}, we 
and by letting the analytical convergence upper bound in \eqref{convergence} satisfy the convergence constraint,\footnote{Optimization using upper bound as an approximation has also been adopted in \cite{wang2019adaptive,tran2019federated, chen2020convergence,luo2020cost}. %and resource allocation based literature \cite{tran2019federated, chen2020convergence,yang2019energy}.%In this case, we can see that \textbf{P2} is more constrained than \textbf{P1}, i.e., any feasible solution  of \textbf{P2} is also feasible for~\textbf{P1}.
} the original Problem \textbf{P1} can be approximated as
\begin{equation}
\begin{array}{cl}
\label{obj2}
\!\!\!\!\!\!\textbf{P2:} \  \min_{\mathbf{q}, R} & \!\!\! \sum_{i=1}^{N}\!\left[\left(\sum\nolimits_{j=1}^{i} q_{j}\right)^{\!K}\!\!-\!\left(\sum\nolimits_{j=1}^{i-1} q_{j}\right)^{\!K}\right]  t_{i}R \\
\quad \text { s.t. } &\!\!\! \frac{1}{R}\left(\alpha {\sum_{i=1}^N\frac{p_i^2G_i^{2}}{q_i}} +\beta\right)\le \epsilon,\\%,\\
&\sum_{i=1}^Nq_i=1,\\
&  q_i>0,  \forall i \in \mathcal{N}, \ R \in \mathbb{Z}^{+}. 
\end{array}
\end{equation}%\end{equation}
%where we  we approximate \textbf{P1} as
%Combining with our convergence upper bound \eqref{convergence}, 

Combining with \eqref{convergence}, we can see that Problem \textbf{P2} is more constrained than Problem \textbf{P1}, i.e., any feasible solution  of Problem \textbf{P2}   is also feasible for Problem~\textbf{P1}. 

We further relax $R$ as a continuous variable to theoretically analyze Problem \textbf{P2}. For this relaxed  problem,  %we show remove the convergence constraint and connect it with the objective function. 
suppose ($\mathbf{q}^*$,  $R^*$) is the optimal solution, then we must have 
\begin{equation} \label{equality_hold}
    \begin{array}{c}
        \frac{1}{R^*}\left(\alpha {\sum_{i=1}^N\frac{p_i^2G_i^{2}}{q_i^*}} +\beta\right)= \epsilon.
    \end{array}
\end{equation}
This is because if  \eqref{equality_hold} holds with an inequality, we can always find an $R^{\prime}< R^*$ that  
%if any feasible solution $\mathbf{q}$ and $R^\prime$ that satisfies the $\epsilon$-constraint in \eqref{obj2} with inequality, we can always decrease this $R^\prime$ to some $R^{\prime\prime} < R^\prime$ which 
satisfies \eqref{equality_hold} with equality, but the solution ($q^*$, $R^{\prime}$) can further reduce the objective function value. Therefore, for the optimal $R$, \eqref{equality_hold} always holds, and we can obtain $R$ from \eqref{equality_hold}
%\begin{equation} \label{eq:R_solution}
%R = \frac{1}{\epsilon }\left(\alpha {\sum_{i=1}^N\frac{p_i^2G_i^{2}}{q_i}} +\beta\right).
%\end{equation}
and substitute it %\eqref{eq:R_solution} 
into the objective of Problem \textbf{P2}. Then, the objective of Problem \textbf{P2} is
\begin{equation}
\label{obj22}
    \!\!\!\!\begin{array}{c}
        \left(\sum\limits_{i=1}^{N}\left[\!\left(\sum\nolimits_{j=1}^{i} q_{j}\!\right)^{\!\!K}\!\!-\!\left(\sum\nolimits_{j=1}^{i-1} q_{j}\!\right)^{\!\!K}\right] \! t_{i}\!\right)\!\left(\alpha {\sum\limits_{i=1}^N\frac{p_i^2G_i^{2}}{q_i}} +\beta\right)\!\!,
         %2\left(t_m+Ke_m\right) \dfrac{T_{\epsilon}}{E}+\left(t_p+Ke_t\right) T_{\epsilon}
     \end{array}\!\!\!\!\!
\end{equation}
%yielding a relaxed Problem \textbf{P2$^\prime$}:
%\begin{equation}
% \label{ob22}
%\small
%\begin{array}{cl}
%\!\!\textbf{P2}^\prime\!:  
%\min_{\mathbf{q}}  \!\!& \!\! \left(\!\sum\limits_{i=1}^{N}\!\left[\!\left(\sum\limits_{j=1}^{i} q_{j}\!\right)^{\!\!K}\!\!-\!\left(\sum\limits_{j=1}^{i-1} q_{j}\!\right)^{\!\!K}\right] \! t_{i}\!\right)\!\left(\alpha {\sum\limits_{i=1}^N\dfrac{p_i^2G_i^{2}}{q_i}} +\beta\right) \\
 %2\left(t_m+Ke_m\right) \dfrac{T_{\epsilon}}{E}+\left(t_p+Ke_t\right) T_{\epsilon} \\
%\quad \text {s.t.} & \!\! \sum_{i=1}^Nq_i=1, \ 0\le q_i\le1, \  \forall i \in \mathcal{N}, 
%\end{array}
%\end{equation}
which\footnote{For ease of analysis, we omit  $\epsilon$ as it is a constant multiplied by the entire objective function.} % in the constants $\alpha $ and $\beta$.}  
is only associated with client sampling probabilities $\mathbf{q}$.

%The objective function in 
The objective function \eqref{obj22}, however, is still difficult to optimize because the sampling probabilities $\mathbf{q}$  %$\Expect[T^{(r)}(\mathbf{q})]$ in
is in a polynomial sum with an order $K$. For analytical tractability, we define an approximation of $\Expect[T^{(r)}(\mathbf{q})]$ as
%the approximation is motivated by uniform sampling, where
\begin{equation}
    \label{appro_Tr}
    \tilde{\Expect}[T^{(r)}(\mathbf{q})]:=\sum_{i=1}^Nq_it_i. %\cdot M(K),
\end{equation}
%where $M(K)$ is an increasing function of $K$, with $M(1)=1$ we 
The approximation $\tilde{\Expect}[T^{(r)}(\mathbf{q})]$ is exactly the same as $\Expect[T^{(r)}(\mathbf{q})]$ in the following two cases.

\noindent\emph{Case 1}: For homogeneous $t_{i}$ ($t_i=t_0, \forall i \in \mathcal{N}$),  we have %$\tilde{\Expect}[t_\textnormal{tot}]=\Expect[t_\textnormal{tot}]$ as
\begin{equation}
\label{avgtimehomo}
\begin{array}{cl}
\!\!\!\!\!\Expect[T^{(r)}(\mathbf{q})]\!\!\!\!\!\!&=\sum_{i=1}^{N}\!\left[\left(\sum\nolimits_{j=1}^{i} q_{j}\right)^{\!K}\!-\left(\sum\nolimits_{j=1}^{i-1} q_{j}\right)^{\!K}\right] t_0%\frac{t_{K}\!+\!C_{K}^{K\!-\!1} t_{K\!+\!1}\!%+\!C_{K+1}^{K-1} t_{K\!+\!2} 
%+\ldots\!+\!C_{N-1}^{K-1} t_{N}}{C_{N}^{K}}
\\
%&=\left( \frac{\sum_{i=1}^{N} t_{i}^{(p)}E+\sum_{i=1}^{N} t_{i}^{(m)}}{N}\right)\frac{T}{E}\\
&=\left[\left(\sum\nolimits_{j=1}^{N} q_{j}\right)^{\!K}\!-0%\left(\sum\nolimits_{j=1}^{0} q_{j}\right)^{\!K}
\right]  t_0
=\left[1^K\!-\!0^K\right]\!t_0\!\\&=\!\sum_{i=1}^Nq_it_i=\tilde{\Expect}[T^{(r)}(\mathbf{q})].
\end{array}
\end{equation}

\noindent\emph{Case 2}: For heterogeneous $t_{i}$ with
%we show that 
 $K=1$, we have %$\tilde{\Expect}[t_\textnormal{tot}]=\Expect[t_\textnormal{tot}]$ as
\begin{equation}
\label{avgtimek=1}
\begin{array}{cl}
\!\!\!\Expect[T^{(r)}(\mathbf{q})]\!\!\!\!\!\!&=\sum_{i=1}^{N}\!\left[\left(\sum\nolimits_{j=1}^{i} q_{j}\right)^{\!K}\!-\left(\sum\nolimits_{j=1}^{i-1} q_{j}\right)^{\!K}\right]  t_i\\
&=\sum_{i=1}^Nq_it_i=\tilde{\Expect}[T^{(r)}(\mathbf{q})].
\end{array}
\end{equation}

For the general case, we can consider $\tilde{\Expect}[T^{(r)}(\mathbf{q})]$ as an approximation to 
$\Expect[T^{(r)}(\mathbf{q})]$. 
Using this approximation, Problem 
\textbf{P2} can be expressed as  
\begin{equation}
\label{ob3}
\begin{array}{cl}
 \!\!\!\! \!\!\textbf{P3:} \  
\min_{\mathbf{q}}  & \left(\sum_{i=1}^Nq_it_i\right)\left(\alpha {\sum_{i=1}^N\dfrac{p_i^2G_i^{2}}{q_i}} +\beta\right) \\%[6pt]
 %2\left(t_m+Ke_m\right) \dfrac{T_{\epsilon}}{E}+\left(t_p+Ke_t\right) T_{\epsilon} \\
\ \ \text {s.t.} & \!\!\!\! \sum_{i=1}^Nq_i=1,\  q_i>0,  \forall i \in \{1,\ldots N\}. 
\end{array}
\end{equation}
\textbf{Remark:} %\textbf{P3} is an approximation of \textbf{P2} due to the use of $\tilde{\Expect}[T^{(r)}(\mathbf{q})]$ in the original objective $\Expect[T_\textnormal{tot}]$. 
The objective function of Problem \textbf{P3} is in a more straightforward form compared to Problem \textbf{P2}. However, to solve for the optimal sampling probabilities $\mathbf{q}$, we need to know the value of the parameters in \eqref{ob3}, e.g., $G_i$, $\alpha$, and $\beta$.\footnote{We assume that clients' heterogeneous time $t_i$ and their dataset size $p_i$ can be measured offline.} 

In the following, we %show how to learn the unknown parameters and 
solve Problem \textbf{P3} as an approximation of the original Problem \textbf{P1}. 
Our empirical results in Section~\ref{sec:experimentation} demonstrate that the solution obtained from solving Problem \textbf{P3} %$\tilde{\Expect}[t_\textnormal{tot}]$ 
achieves superior total wall-clock time performances compared to baseline client sampling schemes. %\emph{near-optimal performance} of the original Problem \textbf{P1}. % $\Expect[t_\textnormal{tot}]$}. 
%For ease of analysis, we incorporate $\epsilon$ and $E$ in the constants $\alpha $ and $\beta$ next.

\subsection{Solving Problem  \textbf{P3}}
%Before we start this subsection, we first emphasize the key challenge to solve Problem  \textbf{P3}. % as well as Problem  \textbf{P2}
Problem  \textbf{P3} is challenging to solve because we can only obtain the unknown parameters $G_i$, $\alpha$ and $\beta$ during the training process of FL. In this subsection, we first show how to estimate these unknown parameters. %via a substitute-sampling method, summarized in Algorithm~2.
Then, we develop an efficient algorithm to  solve Problem \textbf{P3}. We summarize the overall algorithm in  Algorithm~\ref{opt_algorithm}. 
Finally, we identify some insightful solution properties. 

\subsubsection{\textbf{Estimation of Parameters  $G_i$ and $\frac{\alpha }{\beta}$}}
We first show how to estimate $\frac{\alpha }{\beta}$ via a substitute sampling scheme.\footnote{We only need to estimate the value of $\frac{\alpha }{\beta}$ instead of $\alpha$ and $\beta$ each, because we can divide parameter $\alpha$ on the objective of Problem \textbf{P3} without affecting the optimal sapling solution.}  Then, we show that we can indirectly acquire the knowledge of $G_i$ during the estimation process of $\frac{\alpha }{\beta}$. 

%The basic idea is to 
%In fact, we only need to estimate the value of $\frac{\beta}{\alpha }$ instead of $\alpha $ and $\beta$ each. However, the value of $G_i$ is challenging to obtain because we need to know the upper bound of every client's gradient norm throughout the learning process, which is infeasible in practice when $N$ is large. 
%In the following, we propose an efficient sampling-based online algorithm to estimate $G_i$ and $\frac{\alpha}{\beta}$, and show that the estimation cost is marginal. 

The basic idea is to utilize the derived convergence upper bound in \eqref{convergence} to  approximately solve $\frac{\alpha}{\beta}$ as a single variable,  
%Specifically, we run Algorithm 1 in parallel with two  %schemes in parallel 
via performing Algorithm~1 with two
baseline sampling schemes: uniform sampling $\mathbf{q_1}$ with ${q}_i=\frac{1}{N}$ and  weighted sampling $\mathbf{q_2}$ with ${q}_i=p_i$, respectively. 

Note that we only let sampling schemes $\mathbf{q_1}$ and $\mathbf{q_2}$ run  until a  %relative high 
pre-defined loss $F_s$ is reached (instead of running all the way until they converge to the  precision $\epsilon$),
because our goal is to find and run with the optimal sampling scheme $\mathbf{q^*}$ so that we can achieve the target precision with the minimum wall-clock time. % and run with it to achieve the do this  

%Instead of running the  %The key of our method is to let the two schemes run all the way, we  
%and utilize the derived convergence upper bound in \eqref{convergence} to approximate  $\Expect[F\left(\mathbf{w}(\mathbf{q},R)\right)]-F^{*}$. 

Specifically, suppose $R_{\mathbf{q_1},s}$ and $R_{\mathbf{q_2},s}$ are the number of rounds for reaching the pre-defined loss $F_s$ for schemes $\mathbf{q_1}$ and $\mathbf{q_2}$, respectively. {Considering that the training loss decreases quickly at the beginning and slowly when approaching convergence, the values of $R_{\mathbf{q_1}}$ and $R_{\mathbf{q_2}}$ are normally very small compared to the required number of rounds for reaching the target loss (with precision $\epsilon$).} 
According to \eqref{convergence}, we have 
\begin{equation}
\begin{cases}
    \label{A0B01}
   \left(F_s-F^{*}\right)R_{\mathbf{q_1},s} \approx \alpha N{\sum_{i=1}^Np_i^2G_i^{2}} +\beta,\\
      \left(F_s-F^{*}\right)R_{\mathbf{q_2},s} \approx \alpha {\sum_{i=1}^N{p_iG_i^{2}}} +\beta.
 \end{cases}
\end{equation}
Based on \eqref{A0B01}, we have
  \begin{equation}
    \label{A0B02}
    \frac{R_{\mathbf{q_1},s}}{R_{\mathbf{q_2},s}} \approx 
    \frac{ \alpha N{\sum_{i=1}^Np_i^2G_i^{2}} +\beta}{\alpha {\sum_{i=1}^N{p_iG_i^{2}}} +\beta}.
\end{equation}
Then, we can obtain $\frac{\alpha }{\beta}$ from \eqref{A0B02} once we know the value of  $G_i$. 
Notably, we can estimate $G_i$ during the procedure of estimating $\frac{\alpha }{\beta}$. The idea is to let the sampled clients send back the norm of their local SGD along with their returned local model updates, and then the server updates $G_i$ with the received norm values. This approach does not add much communication overhead, since we only need to additionally transmit the value of the gradient norm (e.g., only a few bits for quantization) instead of the full gradient information. {In addition, instead of retraining the model using the calculated $\mathbf{q}^*$ from the initial parameter $\mathbf{w_0}$, we can continue to train the global model after the estimation process, to avoid repeated training and reduce the overall training time.} %which saves the training time for estimation overhead.}

In practice, %as the convergence bound may not be tight, 
due to the sampling variance, we may set several different $F_s$  %and adopt linear regression method 
to obtain an averaged estimation of $\frac{\alpha }{\beta}$. % and a more accurate estimate of $G_i$. 
The overall estimation process corresponds to Lines~\ref{alg:optimalSolution:startEstimation}--\ref{alg:optimalSolution:endEstimation} of Algorithm~\ref{alg:optimalSolution}.

%we run Algorithm 1 with the two schemes in parallel %with an initial model $\mathbf{w}_0\!=\!\mathbf{0}$ 
%until they both reach a pre-defined global losses $F_s$ with executed round numbers $R_{\mathbf{q_1},s}$ and ${R_{\mathbf{q_2}},s}$. Particularly, we 
%additionally let the sampled clients feedback their gradient norm as an estimate of their $G_i$, 
%and we will update their $G_i$ in an online fashion during the estimation process. The pre-defined losses $F_s$ can be set to a relatively high value, to maintain a small estimation cost.} %, but they cannot be too high either as it would cause low estimation accuracy.} %to achieve $\epsilon_1$ precision. 
%Then, substituting sampling scheme $\mathbf{q_1}$ and $\mathbf{q_2}$ into \eqref{convergence}, we have %as adopted in \cite{wang2019adaptive}, we  as 

%Dividing the above two equation, we have %can obtain $\frac{\beta}{\alpha }$ from \eqref{cases} as

%characterize some properties of the  Problem \textbf{P3}. Then, we propose an efficient sampling-based algorithm to learn the convergence-related unknown parameters $\alpha $ and $\beta$, based on which the optimal sampling solution $\mathbf{q}^*$ (of \textbf{P3}) can be efficiently computed. The overall algorithm for obtaining $\mathbf{q}^*$ is given in Algorithm~\ref{alg:optimalSolution}.

\begin{algorithm}[t]
\label{opt_algorithm}
\small
\caption{Approximate Optimal Client Sampling  for FL with System and Statistical Heterogeneity}
\label{alg:optimalSolution}
	\KwIn{{$N$, $K$, $E$, $t_i$, $p_i$,  $\mathbf{w}_0$}, loss  $F_s$,  precision $\epsilon_0$}
	\KwOut{{Approximation of $\mathbf{q}^*$}}
% 	\begin{algorithmic}[1]

\For{$s\leftarrow1,2, \ldots, S$  \label{alg:optimalSolution:startEstimation}}{
Server runs Algorithm 1 with uniform sampling $\mathbf{q_1}$ and weighted sampling $\mathbf{q_2}$, respectively;

The sampled clients send back their local gradient norm information along with their updated models;  

Server updates all clients' $G_i$ based on the received gradient norms;

Server records $R_{\mathbf{q_1},s}$ and $R_{\mathbf{q_2},s}$ when reaching $F_s$;
}

Calculate average $\frac{\alpha}{\beta}$ using \eqref{A0B02}; \label{alg:optimalSolution:endEstimation}

%Choose a feasible starting point $z_0 \leftarrow \left(K_0, E_0\right)$ and set $j \leftarrow 0$; \label{alg:optimalSolution:startOptimization}

\For{$M(\epsilon_0)\leftarrow t_1, t_1+\epsilon_0, t_1+2\epsilon_0 \ldots, t_N$  \label{alg:optimalSolution:startSolve}}{ %$K_{j}-K_{j-1}>\epsilon_0$ or $E_{j}-E_{j-1}>\epsilon_0$}{
Substitute $M(\epsilon_0)$, $\frac{\beta}{\alpha }$, $N$,  $t_i$, $p_i$, $G_i$ into \textbf{P4}; 

Solve \textbf{P4} via CVX, and obtain $\mathbf{q}^*(M(\epsilon_0))$ %$g^*(\mathbf{q}^*(M(\epsilon_0)),M(M(\epsilon_0)))$; %\label{alg:optimalSolution:projectionK}
}

%obtain  

\Return $\mathbf{q}^*(M^*(\epsilon_0))\leftarrow \arg\min_{M(\epsilon_0)}\mathbf{q}^*(M(\epsilon_0))$ %\min_{M(\epsilon_0)}\mathbf{q}^*(M(\epsilon_0))$ %\arg\mathop{\min}\limits_{\mathbf{q}^*(M)}g^*(\mathbf{q}^*(M),M)$ % \leftarrow \left(K_j, E_j\right)$
\label{alg:optimalSolution:endOptimization}

%Numerically solve $E^*$ by substitute $K$ into $\frac{\partial C_\textnormal{tot}}{\partial E}=0$;

%Substitute $E^*$ into  \eqref{opt_K} and derive $K^*$;

% \end{algorithmic}
\end{algorithm}

\subsubsection{\textbf{Optimization Algorithm for  $\mathbf{q}^*$}} %for Problem \textbf{P3}}} 
We first identify the property of Problem \textbf{P3} %with the following theorem, 
and then show how to compute $\mathbf{q}^*$. 
\begin{theorem}
\label{theorem:biconvex}
Problem \textbf{P3} %in \eqref{ob2} 
is non-convex. 
\end{theorem}
\noindent\emph{Proof  Sketch.}
The idea is to show that the Hessian of the objective function in Problem \textbf{P3} is not positive semi-definite. 
{For example, for $N=2$ case, we have
 %\begin{equation}
 %\label{partialq1}
$\frac{\partial^2 \tilde{\Expect}[T_\textnormal{tot}]}{\partial^2 q_1}=\frac{2\alpha q_2t_2p_1^2G_1^2}{q_1^3}>0$, %\end{equation} 
whereas
 %\begin{equation}
  %\label{partialq1q2}
  %\footnotesize
$\frac{\partial^2 \tilde{\Expect}[T_\textnormal{tot}]}{\partial^2 q_1}\frac{\partial^2 \tilde{\Expect}[T_\textnormal{tot}]}{\partial^2 q_2}\!-\!\left(\frac{\partial^2 \tilde{\Expect}[T_\textnormal{tot}]}{\partial q_1\partial q_2}\right)^{\!\!2}\!=\!-\alpha ^2\left(\frac{t_1p_2^2G_2^2}{q_2^2}\!-\!\frac{t_2p_1^2G_1^2}{q_1^2}\right)^{\!\!2}\!\le\!0$.}
%\end{equation} 
  \qedsymbol

\vspace{1mm}
To solve Problem \textbf{P3}, we  define a new control variable % $M$ as
\begin{equation}
    \label{new_M}
    M:=\sum\nolimits_{i=1}^Nq_it_i,
\end{equation}
where $t_1\le M \le t_N$.
Then, we rewrite Problem \textbf{P3} as 
%$M:=\sum_{i=1}^Nq_it_i$, with $t_1\le M \le t_N$, and for any given value of $M\in[t_1,t_N]$, Problem \textbf{P3} can be transformed into a new convex  Problem \textbf{P4} as follows.
\begin{equation}
\label{ob4}
\begin{array}{cl}
 \!\!\!\! \!\!\textbf{P4:} \ \
 \min_{\mathbf{q},M} &g(\mathbf{q},M) := M \cdot \left(\alpha {\sum_{i=1}^N\frac{p_i^2G_i^2}{q_i}} +\beta\right) \\%[6pt]
 %2\left(t_m+Ke_m\right) \dfrac{T_{\epsilon}}{E}+\left(t_p+Ke_t\right) T_{\epsilon} \\
\ \ \text {s.t.} &  \sum_{i=1}^Nq_i=1, \\ &\sum_{i=1}^Nq_it_i=M,\\
&  q_i>0,  \forall i \in \mathcal{N}.
\end{array}
\end{equation}
For any fixed feasible value of $M \in [t_1, t_N]$, Problem \textbf{P4} is convex with $\mathbf{q}$, because the objective function is strictly convex and the constraints are linear. 

We will solve Problem \textbf{P4} in two steps. First, for any fixed $M$, we will solve for the optimal  $\mathbf{q}^*(M)$ in Problem \textbf{P4}, % for any fixed value of $M$,
via a convex optimization tool, e.g., CVX \cite{boyd2004convex}.  This allows us to write the objective function of Problem P4 as $g(\mathbf{q}^\ast(M), M)$. Then we will further solve the problem by using 
a linear search method  with a fixed step-size $\epsilon_0$ over the interval $[t_1, t_N]$, where we use the optimal $M^*(\epsilon_0)$ and the corresponding $\mathbf{q}^*(M^*(\epsilon_0))$ in the search domain to approximate the optimal $M^*$ and  $\mathbf{q}^*$ in Problem \textbf{P4}.  %for the minimum value $g(\mathbf{q}^\ast(M), M)$.
%to find the minimum objective value $g^*(\mathbf{q^*},M^*)$  and the corresponding optimal $\mathbf{q}^*$. 
This optimization process
corresponds to Lines \ref{alg:optimalSolution:startSolve}--\ref{alg:optimalSolution:endOptimization} of Algorithm~2. %Therefore, 

%Therefore, via any convex optimization tool, e.g., CVX, we could obtain the corresponding optimal sampling solution  $\mathbf{q}^*(M)$ and minimum $\tilde{\Expect}[T_\textnormal{tot}^*(M)]$ under the fixed value of $M$.  Then, since $M$ is closed bounded by $[t_1, t_N]$, we could simply adopt the line-search method for $M$ and achieve the global optimal sampling solution $\mathbf{q}^*$ for the non-convex Problem \textbf{P3}. %The complexity for line search is $\frac{1}{\epsilon_0}$ for $\epsilon_0$ precision.  
\textbf{Remark}: Our optimization algorithm is efficient in the sense that the linear search domain $[t_1, t_N]$ is independent of the scale of the problem, e.g., number of $N$. %The effectiveness of our algorithm is  that  

%\textbf{Optimization Algorithms:}
%\textbf{1.} Although  \textbf{P3} is non-convex, since the objective is continuous with respect to $p_i$, we could use Projected Gradient Decent method to achieve a local optimal. Nevertheless,  we cannot tell whether it is global optimum.

%\subsubsection{\textbf{Online Algorithm}}

\subsubsection{\!\textbf{Property of Optimal Client Sampling}}\!\! 
%Besides achieving the optimal sampling solution, 
Next we show some interesting properties of the optimal sampling strategy. %  which characterizes the sampling principle for FL with both system and statistical heterogeneity. %sampling principles by analyzing the  solution properties of Problem \textbf{P3}. 
\begin{theorem}
\label{property1}
Suppose $\mathbf{q}^*$ is the optimal solution of %For any value of $\alpha $ and $\beta$ in 
Problem \textbf{P3}. For two different clients $i$ and $j$, if $t_i\le t_j$ and $p_iG_i\ge p_jG_j$, then $q_i^*\ge q_j^*$. 
\end{theorem}
\noindent\emph{Proof  Sketch.}
The idea is to show by contradiction that if $q_i^*<q_j^*$, we can simply let $q_i^\prime=q_j^*, q_j^\prime=q_i^*$ such that $q_i^\prime>q_j^\prime$ and achieve a smaller $\Expect[T_\textnormal{tot}(q_i^\prime,q_j^\prime)]$ than   $\Expect[T_\textnormal{tot}(q_i^*,q_j^*)]$. 
\  \ \ \qedsymbol
 
\vspace{1mm}
 
Theorem~\ref{property1} shows that the optimal client sampling strategy %under both system and statistical heterogeneity
should allocate higher probabilities to those who have smaller $t_i$ and larger product value of $p_iG_i$, which 
characterizes the impact and interplay between system heterogeneity and statistical heterogeneity. 
Although it may be infeasible to derive an   %closed form solution to identify the exact
analytical relationship regarding the exact impact of $t_i$ and  $p_iG_i$ on $\mathbf{q}^*$ due to non-convexity of Problem \textbf{P3}, we show by Corollary~\ref{property2} that we can obtain the closed-form solution of $\mathbf{q}^*$ with $t_i$ and  $p_iG_i$ when  $\frac{\beta}{\alpha }\rightarrow0$.  
\begin{corollary}
\label{property2}
When $\frac{\beta}{\alpha }\rightarrow0$, 
%If $\alpha \gg \beta$ holds for a certain FL task, we have 
the global optimal solution of Problem \textbf{P3} is
\begin{equation}
    \label{optsampling_statis_sys}
        \begin{array}{c}
            q_{i}^{*}=\frac{p_{i} G_{i}}{\sqrt{t_i}}\left/ \sum_{j=1}^{N} \frac{p_{j} G_{j}}{\sqrt{t_j}}\right. .
        \end{array}
\end{equation}
\end{corollary}
\begin{prf}
If $\frac{\beta}{\alpha }\rightarrow0$, because $\sum\nolimits_{i=1}^Nq_it_i$ is bounded between $\left[t_1, t_N\right]$, we have $\left(\sum\nolimits_{i=1}^Nq_it_i\right)\frac{\beta}{\alpha}\rightarrow 0$. Then, the objective of Problem \textbf{P3} can be written as
\begin{equation}
\label{special_case}
    \begin{array}{c}
      \min_{\mathbf{q}}   \left(\sum\nolimits_{i=1}^Nq_it_i\right)\left( {\sum\nolimits_{i=1}^N\frac{p_i^2G_i^{2}}{q_i}} \right).  
    \end{array}
\end{equation}
By Cauchy-Schwarz inequality, we have %the minimum value of \eqref{special_case} is    
\begin{equation}
    \label{special_case_prove}
    %\begin{array}{cl}
  %  \left(\sum\limits_{i=1}^Nq_it_i\right)\!\left({\sum\limits_{i=1}^N\dfrac{p_i^2G_i^{2}}{q_i}} \right)
  \!\!\begin{array}{c}
   \left(\sum\limits_{i=1}^N\left(\sqrt{q_it_i}\right)^2\right)\!\left({\sum\limits_{i=1}^N\left(\frac{p_iG_i}{\sqrt{q_i}}\right)^{\!2}} \right)\ge %\left(\!\sum\limits_{i=1}^N\!\sqrt{q_it_i}\cdot\frac{p_iG_i}{\sqrt{q_i}}\right)^{\!\!2}\\   &=
   \left(\!\sum\limits_{i=1}^N\sqrt{t_i}\cdot{p_iG_i}\right)^{\!2}.
  \end{array}\!
\end{equation}
Hence, the minimum of \eqref{special_case} is $\left(\sum_{i=1}^N\sqrt{t_i}\cdot{p_iG_i}\right)^{\!2}$, which  %a constant value and 
is independent of $\mathbf{q}$. The equality of \eqref{special_case_prove}  holds if and only if $\sqrt{q_it_i}=c\cdot \frac{p_iG_i}{\sqrt{q_i}}$ (for an arbitrary scalar $c$). Noting $\sum_{i=1}^N q_i = 1$ yields \eqref{optsampling_statis_sys} and concludes the proof. % with the equality holds only when  \eqref{optsampling_statis_sys} holds, which concludes the proof. %Therefore, the minimum $\tilde{\Expect}[T_\textnormal{tot}]$ is independent to 
\end{prf}
Though Corollary~\ref{property2} is valid only for the special case of $\frac{\beta}{\alpha }\rightarrow0$, the global optimal sampling solution in \eqref{optsampling_statis_sys} characterizes an analytical interplay between the system heterogeneity (${t_i}$) and statistical heterogeneity ($p_iG_i$).  %sampling solution of pure statistical heterogeneity in \eqref{optsampling_statis} in Corollary~\ref{opt_q_statis}. 
Particularly, when $t_i\!=\!t_0$, for each $i\! \in \mathcal{N}$, $\mathbf{q}^*$ in \eqref{optsampling_statis_sys} recovers the optimal sampling solution in Corollary~\ref{opt_q_statis} for  homogeneous systems. % statistical heterogeneity. \textcolor{blue}{This is intuitive because when clients are with same communication time, the round  time would be fixed, and  minimizing total learning time is equivalent to minimizing the number of total communication rounds.  In other words,  our proposed sampling scheme in Corollary~\ref{opt_q_statis} is optimal for minimizing wall-clock time for homogeneous system.} 
% When $\frac{\beta}{\alpha }$ is large enough, $t_i$ will play a dominate role for $q_i$.

\begin{figure}[!t]
	\centering
	\includegraphics[width=8.8cm,height=3.4cm]{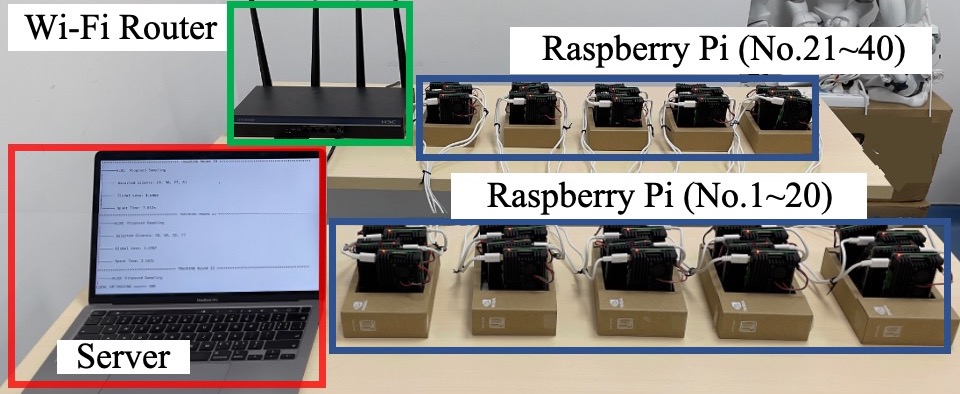}
% 	\vspace{-1mm}
	\caption{Hardware prototype with the laptop serving as the central server and 40 Raspberry Pis % Jetson Nano 
	serving as clients. During the FL experiments, we place the  router  5 meters away from all the devices.}
\label{HDP_new}
% \vspace{-1mm}
\end{figure}

\section{Experimental Evaluation}
\label{sec:experimentation}
In this section, we empirically evaluate the performance of our proposed  client sampling scheme (Algorithm 2) and compare it with four other benchmarks in each round:
%\begin{enumerate}
   % \item 
  1) \emph{full participation},
   % \item 
  2) \emph{uniform sampling},
   % \item 
  3) \emph{weighted sampling}, 
   % \item
  and 4) \emph{statistical sampling} where we sample clients according to Corollary~\ref{opt_q_statis}.
%\end{enumerate}
Benchmarks 1--3 are widely adopted for convergence guarantees in \cite{li2019convergence,haddadpour2019convergence,karimireddy2019scaffold,yang2021achieving, qu2020federated}. The fourth baseline is  an offline variant of the proposed schemes in \cite{chen2020optimal, rizk2020federated}.\footnote{The client sampling in \cite{chen2020optimal, rizk2020federated} is weighted by the norm of the local stochastic gradient in each round, which frequently requires the knowledge of stochastic gradient from all clients to calculate the sampling probabilities.}

%where the gradient norm information are requested in every round to calculate the sampling probability.%, whereas ours but performed in an offline fashion except that  

%The key insight of our experiments is that an optimized client sampling probability can significantly speed up the training time compared to other naive sampling schemes. 
In the following, we first present  the evaluation setup and then show the experimental results.  %We will make our experiment code open-sourced if the paper gets accepted. 

%{We computea high accuracy solution $x^*$ by solving the logistic regressionproblem with minimum loss value found $0.XXX$. Thereafter, the metric used is the difference between the current objective value and the optimal one, i.e. $\Expect[F\left(\mathbf{w}(\mathbf{q},R)\right)]-F^{*}$. In all experiments, the stopping criterion is set such that a certain accuracy is achieved, e.g., $\epsilon=0.001$.  }

%\begin{figure}[!b]
%	\centering
%	\includegraphics[width=8cm,height=1.5cm]{run_time.jpg}
%	\vspace{-2mm}
%	\caption{Measured heterogeneous communication and computation time in one round with $E=100$ and $K=5$ randomly sampled clients.}
%	\label{fig_sim}
%\vspace{-3mm}
%\end{figure}

%\begin{comment}

\begin{table*}[htbp]
\scriptsize
 \caption{Performances of Wall-clock Time for  Reaching Target {Loss} for Different Sampling Schemes}%\vspace{-0.05in}}
 \label{sample3}
  \centering
  \begin{threeparttable}  
    \begin{tabular}{c||c|c|c|c|c}
    \toprule
   \diagbox[]{Setup}{Sampling scheme} &  \textbf{proposed sampling} & statistical sampling & weighted sampling & uniform sampling & full participation \bigstrut\\
    \hline
   {\makecell[c]{\textbf{Prototype Setup} (EMNIST dataset)}} &  $\mathbf{733.2}$ \textbf{s} & $2095.0$ s ($\mathbf{2.9\times}$) & $2221.7$ s ($\mathbf{3.0\times}$) & $2691.5$ s ($\mathbf{3.7\times}$)\tnote{\dag} & $2748.4$ s ($\mathbf{3.7\times}$) \bigstrut\\
%\cline{2-7}           & accuracy: 68\% & $\mathbf{733.2}$ \textbf{s} & $2226.3$ s & $2922.7$ s & $2983.6$ s & $2917.9$ s \bigstrut\\
    \hline
   {\makecell[c]{\textbf{Simulation Setup 1}  (Synthetic  dataset)} }  & $\mathbf{445.5}$ \textbf{s} & $952.4$ s ($\mathbf{2.1\times}$)& $940.2$ s ($\mathbf{2.1\times}$) & $933.8$ s ($\mathbf{2.1\times}$) & $1526.8$ s ($\mathbf{3.4\times}$) \bigstrut\\
%\cline{2-7}           & accuracy: 75.3\% & $\mathbf{445.5}$ \textbf{s} & $774.2$ s & $877.2$ s & $755.4$ s & $1517.5$ s \bigstrut\\
    \hline
{\makecell[c]{\textbf{Simulation Setup 2} (MNIST   dataset)}} &  $\mathbf{245.5}$ \textbf{s} & $373.8$ s ($\mathbf{1.5\times}$) & $542.9$ s ($\mathbf{2.2\times}$) & NA & $898.1$ s ($\mathbf{3.7\times}$) \bigstrut\\
%\cline{2-7}           & accuracy: 96.5\% & $\mathbf{245.5}$ \textbf{s} & $444.2$ s & $688.8$ s & NA     & $794.5$ s \bigstrut\\
    \bottomrule
    \end{tabular}%
     \begin{tablenotes}    
        \scriptsize%
       %\myfont               
        \item \dag \  ``$3.7\times$"  represents the wall-clock time ratio of uniform sampling over proposed sampling  for reaching the target loss, which is equivalent to proposed sampling takes $73$\% less time than uniform sampling.      
      \end{tablenotes}            
    \end{threeparttable}    
  \label{tab1}%
    \vspace{-1mm}
\end{table*}

\begin{figure*}[!t]
\centering
\subfigure[Loss with wall-clock time]{\label{hd_lossa}\includegraphics[width=5.8cm,height=4.3cm]{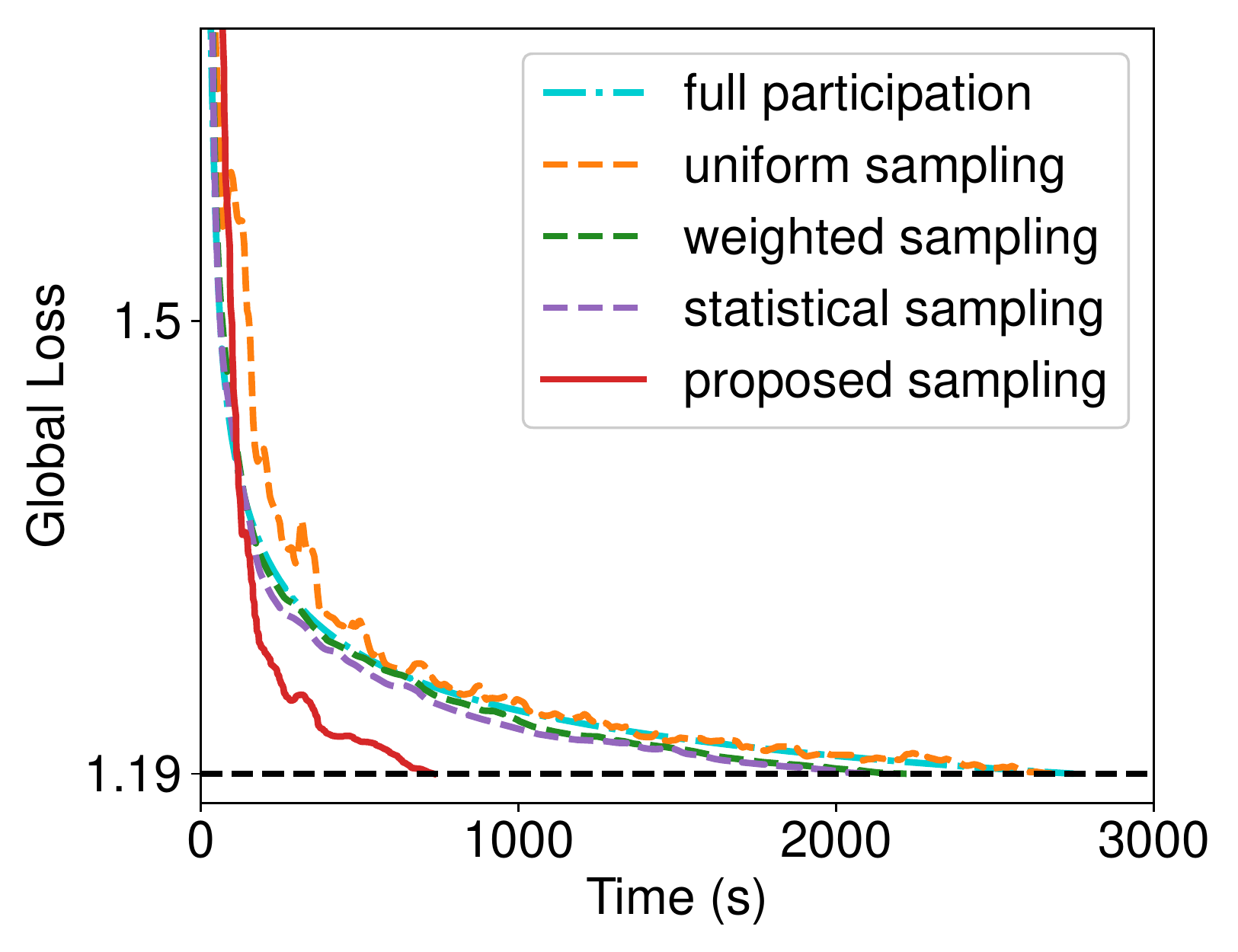}}
\subfigure[Accuracy with wall-clock time]{\label{hd_lossb}\includegraphics[width=5.8cm,height=4.3cm]{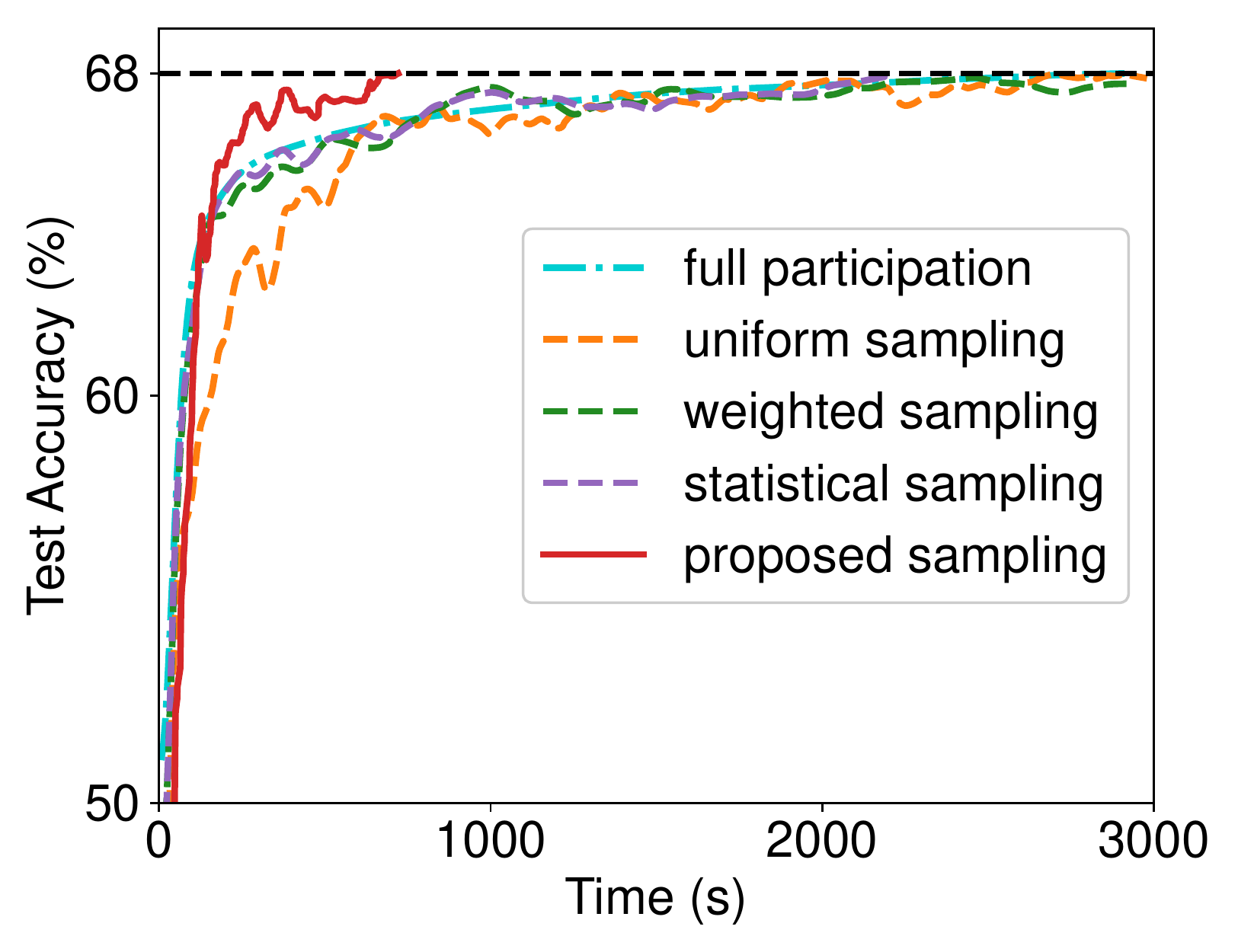}}
\subfigure[Loss with number of rounds]{\label{hd_lossc}\includegraphics[width=5.8cm,height=4.3cm]{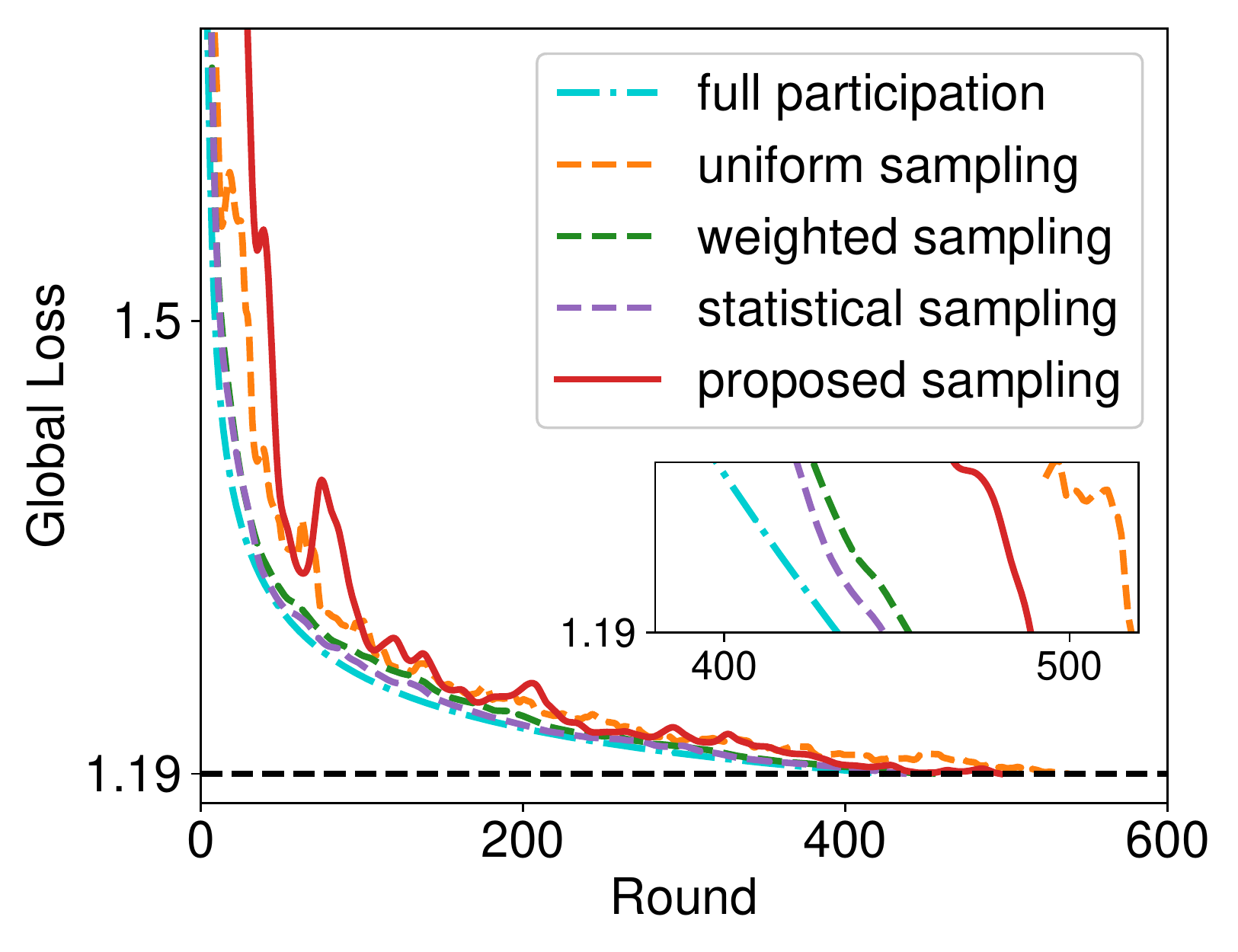}}
\caption{%Performances of $t_\textnormal{tot}$ for different $\left(K, E\right)$ in 
Performance of \textbf{Prototype Setup} with logistic regression model, EMNIST dataset,  uniform communication time, and target loss $1.19$. %$t_i \sim \mathcal{U}(0.187,7.159)$ s.
%(a): Our proposed adaptive sampling achieves the target loss $1.19$ using $714.2$ s compared to weighted sampling $2094.3$ s, statistical sampling $2214.1$ s, full sampling $2729.3$ s, and uniform sampling $2775.7$ s.  (b):  Our  scheme reaches $68$\% test accuracy using $714.2$ s, compared to the other baselines with more than $2700$ s. %weighted sampling $2094.3$ s, statistical sampling $2214.1$ s, full sampling $2729.3$ s, and uniform sampling $2775.7$ s.
%(c): Our scheme reaches the target loss $1.19$ using $491$ rounds, compared to the number of rounds for full sampling $427$, statistical sampling $440$, weighted sampling $451$, and uniform sampling $540$.}
 %A large $E$ may require less round number to achieve communication-efficiency, but can result in a long wall-clock time. 
%(d) Our solution achieves near-optimal performance;  $t_\textnormal{tot}$ first decreases and then increases when $E$ increases, but strictly decreases as $K$ increases.
}
\label{hd}
\vspace{-0.1in}
\end{figure*}

\subsection{Experimental Setup}
\subsubsection{Platforms}
We conduct experiments both on a networked hardware prototype system and in a simulated environment.\footnote{The prototype implementation allows us to capture real system operation time, and the simulation system allows us to simulate large-scale FL environments with manipulative parameters.} 
As illustrated in Fig. \ref{HDP_new}, our prototype system consists of $N=40$ Raspberry Pis serving as clients %and 10 Jetson Nanos, 
and a laptop computer acting as the central server. All devices are interconnected via an enterprise-grade Wi-Fi router. We develop a TCP-based socket interface for the communication between the server and clients with bandwidth control. In the simulated system, we simulate $N=100$ virtual devices and a virtual central server.  

\subsubsection{Datasets and Models} We evaluate our results on two real datasets and a synthetic dataset. For the real dataset, we adopted the widely used MNIST dataset and EMNIST dataset %, which contains 62-class gray-scale images of handwritten digits and characters 
\cite{li2018federated}. %, which contains square $28\times28=784$ pixel gray-scale images of $70,000$ handwritten digits ($60,000$ for training and $10,000$ for testing). 
For the synthetic dataset, we follow a similar setup to that in \cite{li2019convergence}, which generates $60$-dimensional random vectors as input data. %The synthetic data is denoted by $Synthetic \  (\alpha, \beta)$ with $\alpha$ and $\beta$ representing the statistical heterogeneity (i.e., how non-i.i.d. the data are). %how much local models differ from each other and $\beta$ controls how much the local data at each device differs from that of other devices.
%Similar to \cite{li2018federated,li2019convergence}, 
We adopt both the \emph{convex} {multinomial logistic regression} model and the \emph{non-convex}  convolutional
neural network (CNN) model with LeNet-5 architecture \cite{lecun1998gradient}.%2 fully convolutional layers, Relu activation function and MaxPooling. %for both datasets.
%\footnote{The form of the prediction model is  $f\left(\mathbf{w} ; \mathbf{x}_{k}\right)=\operatorname{softmax}\left(\mathbf{W} \mathbf{x}_{k}+\mathbf{b}\right)$  with parameter $\mathbf{w}=(\mathbf{W}, \mathbf{b})$.%For MNIST dataset the input size is $x \in \mathbb{R}^{784}$, and the model size is  $\mathbf{W}_{k} \in \mathbb{R}^{10 \times 784}$ and $\mathbf{b}_{k} \in \mathbb{R}^{10}$. For synthetic dataset, the input size is $x \in \mathbb{R}^{60}, \mathbf{W}_{k} \in \mathbb{R}^{10 \times 60}$ and $\mathbf{b}_{k} \in \mathbb{R}^{10}$.} %which is is a stochastic  optimization problem. We also conduct 
%and the non-convex {deep convolutional neural network (CNN)} model with LeNet-5 architecture \cite{lecun1998gradient}.

\subsubsection{Implementation}
we consider three experimental setups.
\begin{itemize}
    \item \textbf{Prototype Setup}: We conduct the first experiment on the prototype system using  logistic regression and the EMNIST dataset. To generate heterogeneous data partition, similar to \cite{li2018federated}, we randomly subsample $33,036$ lower case character samples from the EMNIST dataset and distribute  among $N\!=\!40$ edge devices in an \emph{unbalanced} (i.e., different devices have different numbers of data samples, following a  power-law distribution) and \emph{non-i.i.d.} fashion (i.e., each device has a randomly chosen number of classes, ranging from $1$ to $10$).\footnote{The number of samples and the number of classes are randomly matched, such that clients with more data samples may not have more classes.} %: the amount of data are distributed in power-law  with each device containing a balanced number of $300$ samples of only $2$ digits labels. % We focus on analyzing $t_\textnormal{tot}$ in this subsection, as 
    \item   \textbf{Simulation Setup 1}: We conduct the second experiment in the simulated system using  logistic regression and the Synthetic dataset. To simulate a heterogeneous setting, we use the non-i.i.d.  $Synthetic \ (1, 1)$ setting. We generate $20,509$ data samples and distribute them among $N\!=\!100$ clients in an \emph{unbalanced} power-law distribution.%,  where the number of samples in each device has a mean of $245$ and standard deviation of $362$. 

     \item \textbf{Simulation Setup 2}: We conduct the third experiment in the simulated
system using CNN and the MNIST dataset, where we randomly subsample 
$15,129$ data samples from MNIST and distribute them among $N \!=\! 100$ clients in an \emph{unbalanced} (following the power-law distribution) and \emph{non-i.i.d.}  (i.e., each device has 1--6 classes) fashion.%\footnote{The number of samples and the number of classes are randomly matched, such that clients with more data samples may not have more classes.}
     
\end{itemize}

\subsubsection{Training  Parameters}  For all experiments, we initialize our model with $\mathbf{w}_0\!=\!\mathbf{0}$ and use an SGD batch size of $b=24$. We use an initial learning rate of $\eta_0 =0.1$ with a decay rate of $\frac{{\eta}_0}{1+r}$, where $r$ is the communication round index. We adopt the similar FedAvg settings as in \cite{bonawitz2019towards,nishio2019client,chen2020optimal,li2019convergence}, where we sample $10$\% of all clients in each round, i.e., $K\!=\!4$ for Prototype Setup and $K\!=\!10$ for Simulation Setups, with each client performing $E=50$ local iterations.\footnote{We also conduct experiments both on Prototype and Simulation Setups with variant $E$ and $K$,  which show a similar performance as the experiments in this paper, and due to page limitations, we do not illustrate them all.} 
%For Prototype Setup, we sample $K\!=\!4$ clients from totally $N\!=\!40$ devices in each round, with each sampled client perform $E\!=\!50$ local iterations. For Simulation Setup, we sample $K\!=\!10$ clients from totally $N\!=\!100$ devices in each round, with each sampled client perform $E\!=\!100$ local iterations.

%uniformly sample $K$ devices at random, which run $E$ steps of SGD in parallel. % For the prototype system and the simulation system, we use decay rate $\frac{{\eta}_0}{1+r}$, where $\eta_0 =0.1$ and $r$ is communication round index. 
%We evaluate the aggregated model in each round on the global loss function. %For fair comparison, %we control all randomness in experiments so that the set of select devices is the same across all different configurations, and 

% We set batch size $b=64$ for both synthetic datasets and MNIST dataset.
\subsubsection{Heterogeneous System Parameters}
For the Prototype Setup, %The prototype system allows us to capture real system operation delay and variance. Specifically,
to enable a heterogeneous communication time, we control clients' communication bandwidth and generate a uniform distribution $t_i \sim \mathcal{U}(0.187,7.159)$ seconds, with a mean of $3.648$ seconds and the standard deviation of $2.071$ seconds.
%we allocate ifferent devices with  measured the average  $t_i\!=\!3.1 \times 10^{-3}$s with standard deviation $2.3 \times 10^{-4}$s. %as illustrated in Fig.~\ref{fig_sim}.
%We do not capture the energy cost in the prototype system because it is difficult to measure.
For the simulation system, we generate  the client transmission delays with an exponential distribution, i.e., $t_{i} \!\sim\! \exp{(1)}$ seconds, with  
both mean and standard deviation as $1$ second.% and variance $1$ s. %Nevertheless, 
%The metric preference can be reflected by the value of trade-off factor $\gamma$. 
% For synthetic datasets, we use batch size $b=64$ and target loss value 0.50 compared to the minimum loss value found 0.3478 (equivalent  to precision $\epsilon=0.01522$). 
 %For MNIST dataset, we use batch size $b=64$ and target loss value 0.50 compared to the minimum loss value found 0.34 (equivalent  to precision $\epsilon=0.X$).
% We evaluate the aggregated model each round on the global loss function. %For fair comparison, %we control all randomness in experiments so that the set of select devices is the same across all different configurations, and 
 %Each evaluated result is averaged over 50 times experiments.
%for randomness error during sampling. 

% \usepackage{booktabs}
% \usepackage{multirow}

\begin{figure*}[!t]
\centering
\subfigure[Loss with wall-clock time]{\label{soft_loss}\includegraphics[width=5.8cm,height=4.3cm]{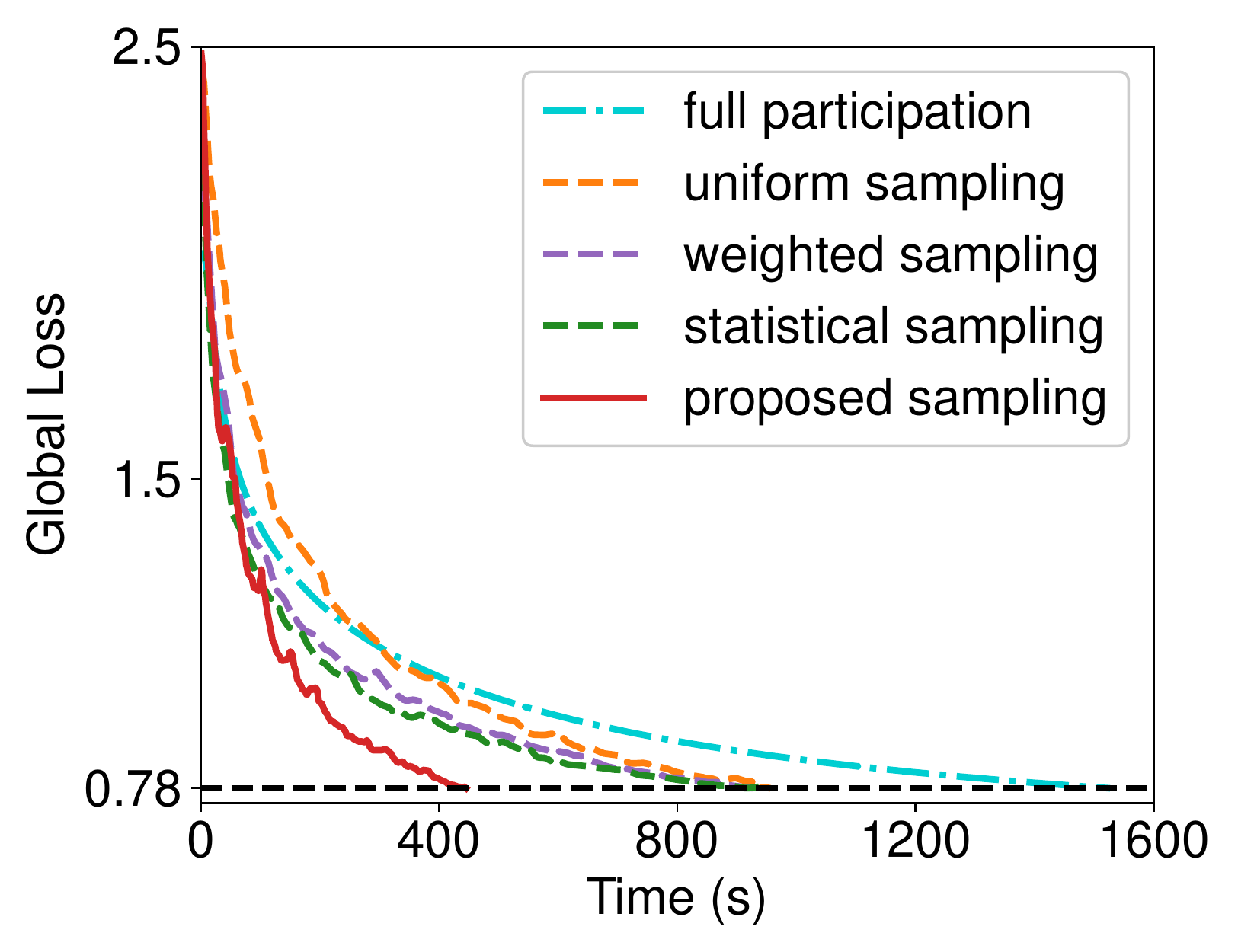}}
\subfigure[Accuracy with wall-clock time]{\label{soft_acc}\includegraphics[width=5.8cm,height=4.3cm]{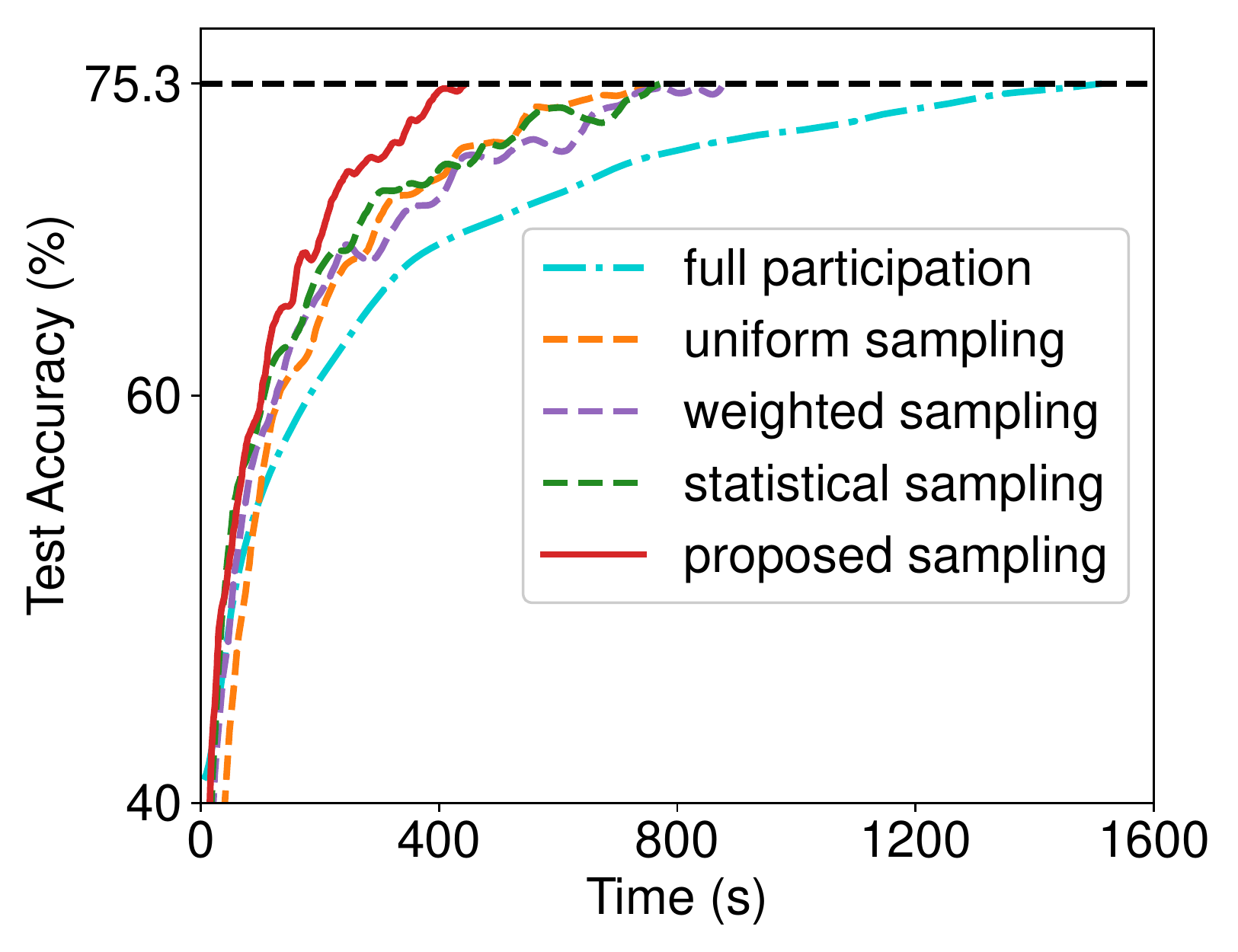}}
\subfigure[Loss with number of rounds]{\label{soft_loss_round}\includegraphics[width=5.8cm,height=4.3cm]{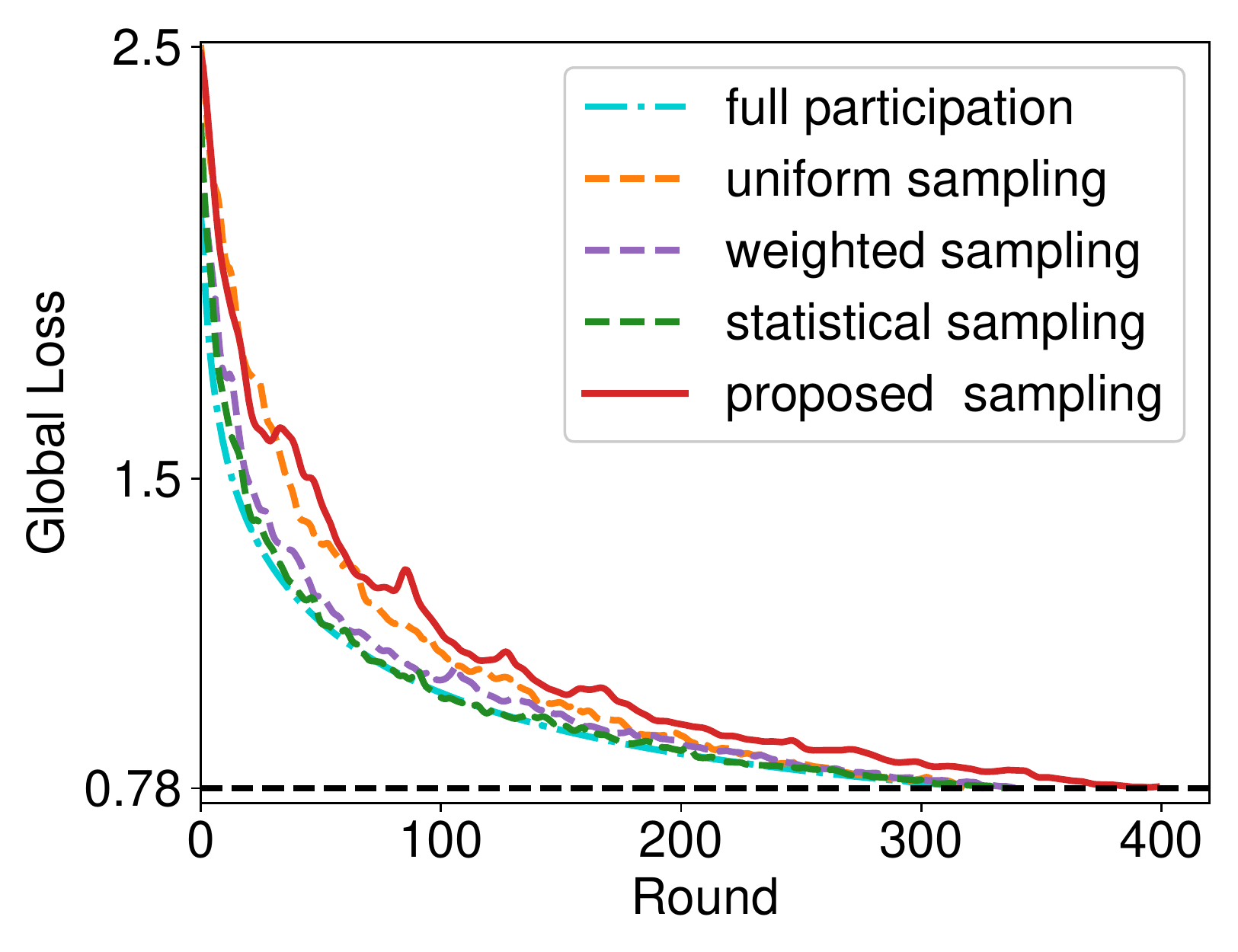}}
\caption{%Performances of $t_\textnormal{tot}$ for different $\left(K, E\right)$ in 
Performance of \textbf{Simulation Setup 1} with logistic regression model, $Synthetic \ (1, 1)$ dataset,   exponential communication time, and target loss $0.78$. % $t_{i} \sim \exp{(1 \text{s})}$. 
%(a): Our proposed scheme achieves the target loss $0.78$ using $433.6$ s compared to statistical sampling $928.0$ s, weighted sampling $931.2$ s, uniform sampling $933.1$ s, and full sampling $1478.5$ s.  (b):  Our proposed scheme reaches $75.3$\% test accuracy using $433.6$ s, compared to full participation around $1460$ s and other sampling schemes around $800$ s. %weighted sampling $2094.3$ s, statistical sampling $2214.1$ s, full sampling $2729.3$ s, and uniform sampling $2775.7$ s.
%(c): Our scheme reaches the target loss $1.19$ using $399$ rounds, compared to other baseline schemes with around $330$ rounds.
}
 %A large $E$ may require less round number to achieve communication-efficiency, but can result in a long wall-clock time. 
%(d) Our solution achieves near-optimal performance;  $t_\textnormal{tot}$ first decreases and then increases when $E$ increases, but strictly decreases as $K$ increases.}
\label{soft}
\vspace{-0.01in}
\end{figure*}

\begin{figure*}[!t]
\centering
\subfigure[Loss with wall-clock time]{\label{soft_loss_2}\includegraphics[width=5.8cm,height=4.3cm]{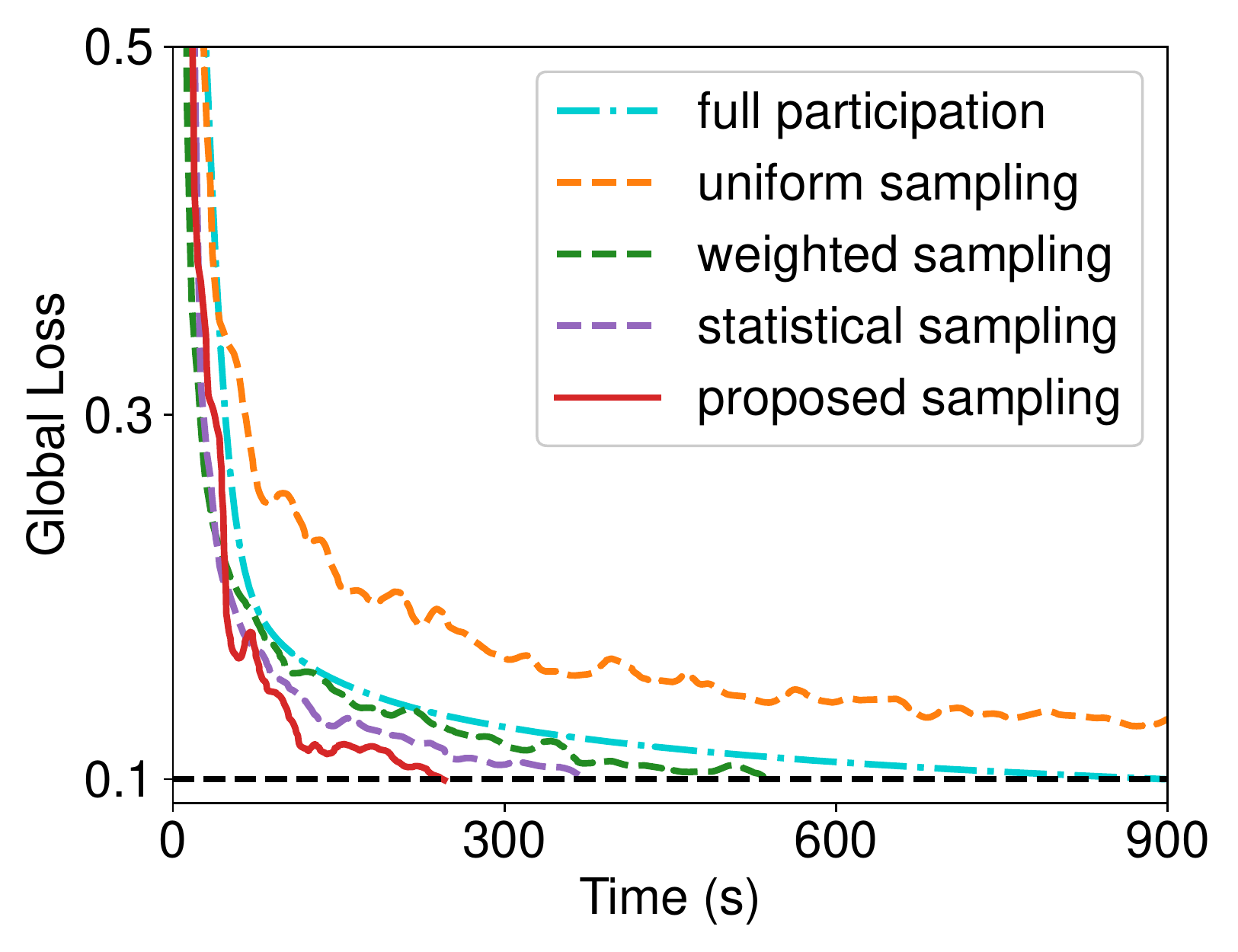}}
\subfigure[Accuracy with wall-clock time]{\label{soft_acc_2}\includegraphics[width=5.8cm,height=4.3cm]{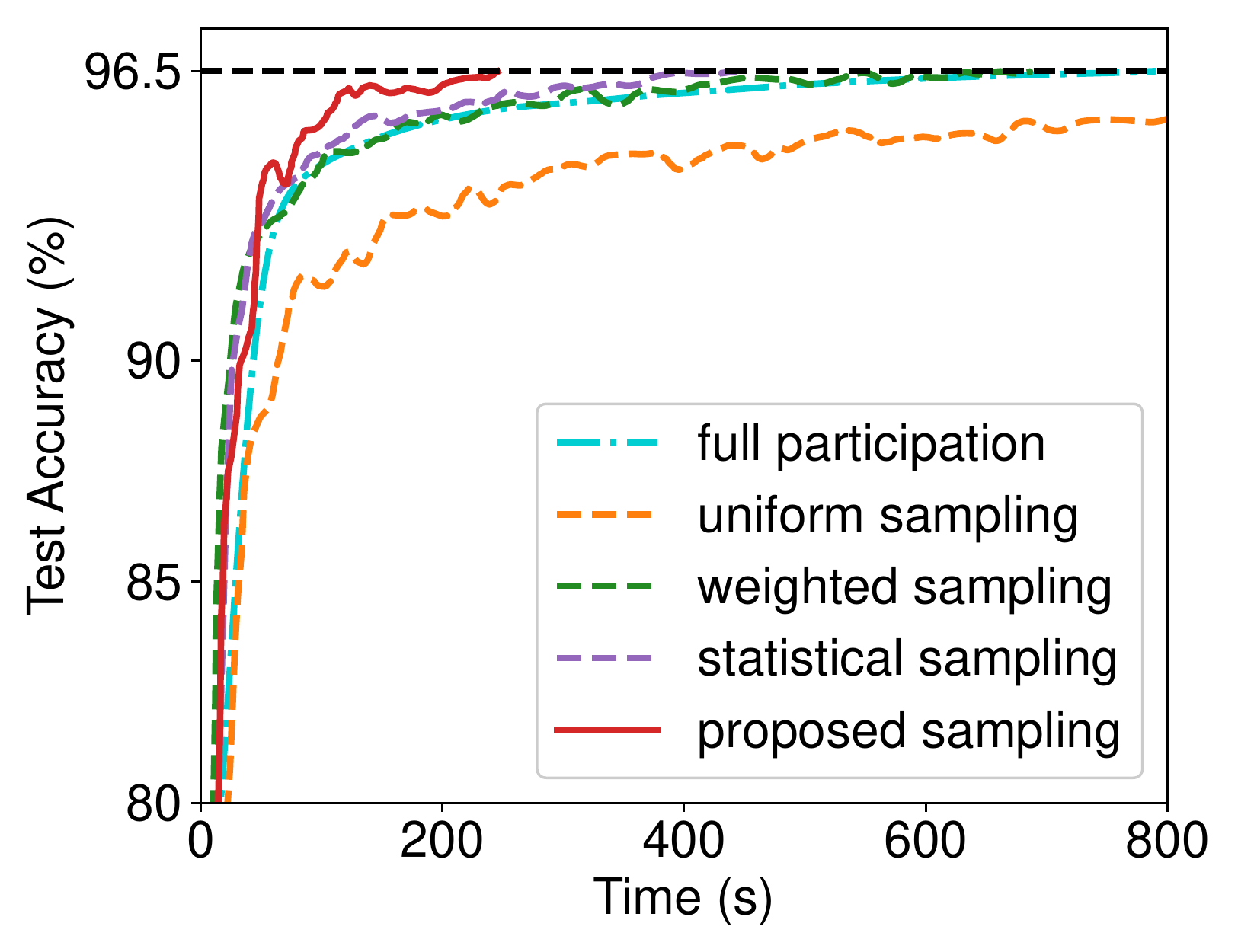}}
\subfigure[Loss with number of rounds]{\label{soft_loss_round_2}\includegraphics[width=5.8cm,height=4.3cm]{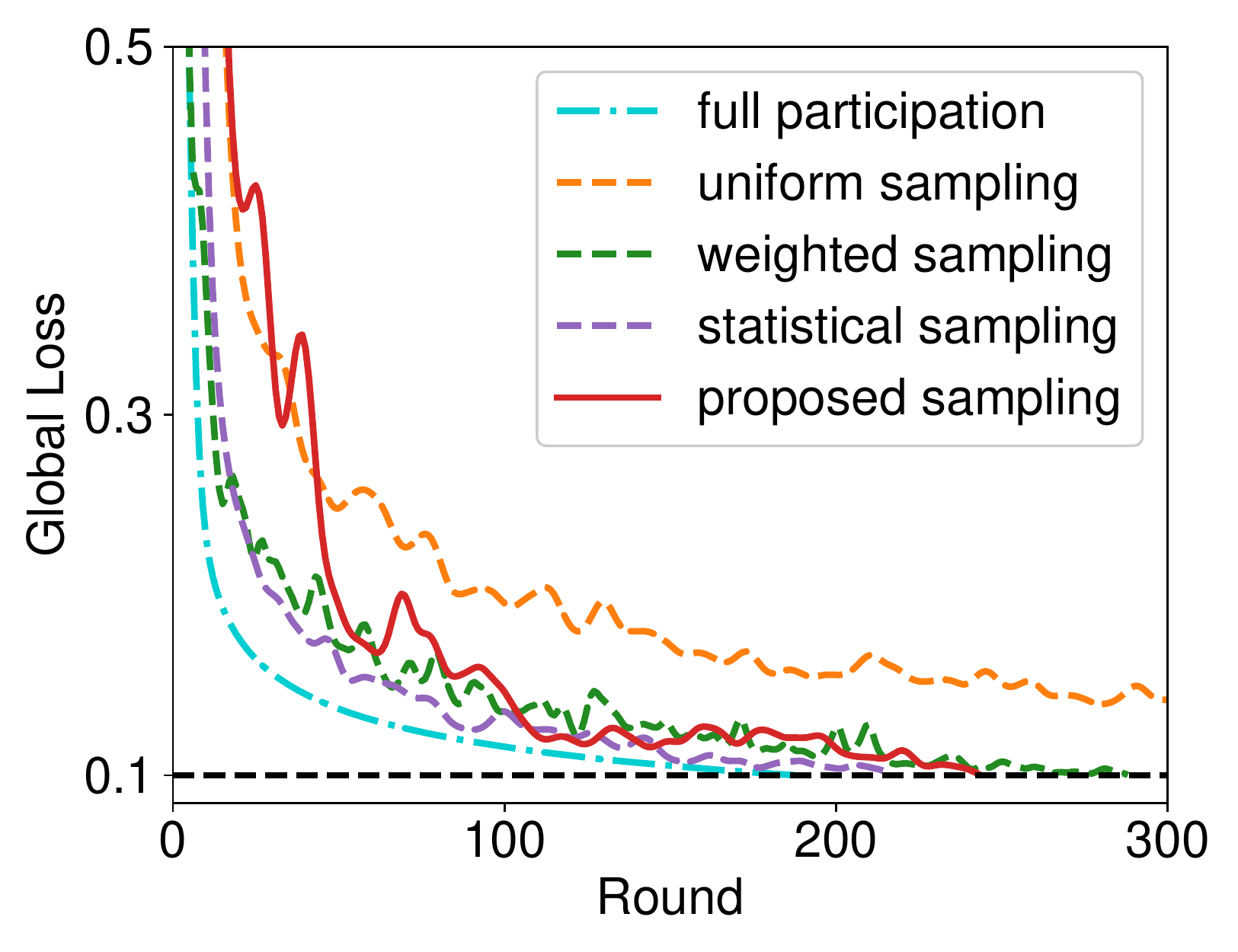}}
\caption{%Performances of $t_\textnormal{tot}$ for different $\left(K, E\right)$ in 
Performance of \textbf{Simulation Setup 2} with CNN model, MNIST dataset,   exponential communication time, and target loss $0.1$. % $t_{i} \sim \exp{(1 \text{s})}$. 
%(a): Our proposed scheme achieves the target loss $0.78$ using $433.6$ s compared to statistical sampling $928.0$ s, weighted sampling $931.2$ s, uniform sampling $933.1$ s, and full sampling $1478.5$ s.  (b):  Our proposed scheme reaches $75.3$\% test accuracy using $433.6$ s, compared to full participation around $1460$ s and other sampling schemes around $800$ s. %weighted sampling $2094.3$ s, statistical sampling $2214.1$ s, full sampling $2729.3$ s, and uniform sampling $2775.7$ s.
%(c): Our scheme reaches the target loss $1.19$ using $399$ rounds, compared to other baseline schemes with around $330$ rounds.
}
 %A large $E$ may require less round number to achieve communication-efficiency, but can result in a long wall-clock time. 
%(d) Our solution achieves near-optimal performance;  $t_\textnormal{tot}$ first decreases and then increases when $E$ increases, but strictly decreases as $K$ increases.}
\label{soft2}
\vspace{-0.1in}
\end{figure*}

\subsection{Performance Results}
% This section shows the main results of our three experiments. 
We evaluate the wall-clock time performances of both the  global training loss and test accuracy on the aggregated model in each round for all sampling schemes. 
%{We computea high accuracy solution $x^*$ by solving the logistic regressionproblem with minimum loss value found $0.XXX$. Thereafter, the metric used is the difference between the current objective value and the optimal one, i.e. $\Expect[F\left(\mathbf{w}(\mathbf{q},R)\right)]-F^{*}$. In all experiments, the stopping criterion is set such that a certain accuracy is achieved, e.g., $\epsilon=0.001$.  }
We average each experiment over $50$ independent runs. For a fair comparison, we use the same random seed to compare sampling schemes in a single run and vary random seeds across different runs.

Fig.~\ref{hd}--\ref{soft2} show the results of Prototype Setup, Simulation Setup 1, and  Simulation Setup 2, respectively. We summarize the key observations as follows.  
%We first show the performances of wall-clock time for different sampling schemes. Then, we show compare the convergence speed with respect to training rounds and wall-clock time. 

\subsubsection{Loss with Wall-clock Time}
As predicted by our theory, Figs.~\ref{hd}(a)--%, Fig.~\ref{soft}(a), and 
\ref{soft2}(a) show that \emph{our proposed sampling scheme achieves the same target loss with significantly less time}, compared to the baseline sampling schemes. Specifically, for Prototype Setup in Fig.~\ref{hd}(a), our proposed sampling scheme spends around $73$\% less time  %runs almost 4$\times$ faster 
than full sampling and uniform sampling and around $66$\% less time than weighted sampling and statistical sampling for  reaching  the  same  target  loss. Fig.~\ref{soft2}(a) highlights the fact that  our
proposed sampling works well with the non-convex CNN model, under which the  naive uniform sampling cannot reach the target loss within $900$ seconds, indicating the importance of a careful client sampling design.   
Table~\ref{tab1} summarizes the superior performances of our proposed sampling scheme in wall-clock time for reaching target {loss} in all three setups. % summarizes our proposed scheme's superior performances of wall-clock time for training loss in all three settings. 
%3$\times$ faster than the other two sampling schemes. % %We also show the superior performances of loss for equalized wall-clock time of our sampling schemes over the other baselines in Table~I. 

\subsubsection{Accuracy with Wall-clock Time}
As shown in Fig.~\ref{hd}(b)--%, Fig.~\ref{soft}(b), and Fig.~
\ref{soft2}(b), our proposed sampling scheme \emph{achieves the target test accuracy\footnote{In Fig.~\ref{hd}(b), Fig.~\ref{soft}(b), and Fig.~\ref{soft2}(b), the target test accuracy corresponds to the test accuracy result when our proposed scheme reaches the target loss. %is set to be the test accuracy result when our proposed sampling scheme reaches the target loss so that the loss and accuracy of our proposed sampling scheme are measured at the same time.
} much faster than 
the other benchmarks}. % cost much longer time for achieving the same test accuracy}. 
Notably, for Simulation Setup 1 with the target test accuracy of $75.3$\% in Fig.~\ref{soft}(b), our proposed sampling scheme takes around $70$\% less time than full sampling and around $46$\% less time than the other sampling schemes. We can also observe the superior test accuracy performance of our proposed sampling schemes in Prototype Setup and non-convex Simulation Setup $2$ in Fig.~\ref{hd}(b) and Fig.~\ref{soft2}(b), respectively.  
%runs almost 3.5$\times$ faster than full sampling, and 2$\times$ faster than the rest of sampling schemes. %Table~1 summarizes the performances of test accuracy for equalized wall-clock time for different sampling schemes. %
%Table~II shows the superior performances of test accuracy  of our sampling schemes of our sampling scheme  over other baselines for equalized time budget. 

\subsubsection{Loss with Number of Rounds}
Fig.~\ref{hd}(c)--%, Fig.~\ref{soft}(c), and Fig.~
\ref{soft2}(c) show that  our proposed sampling scheme requires more training rounds for reaching the target loss compared to baseline statistical sampling and full participation schemes. This observation is expected since our proposed sampling scheme \emph{aims to minimize the wall-clock time instead of the number of rounds.} %In other words, our sampling scheme balances the system and statistical heterogeneity via choosing clients with valuable data (both data quantity and quality) as well as fast communication rate.
Nevertheless, we notice that statistical sampling %has a similar performance with full client participation, but 
performs better than the other sampling schemes, which verifies  Corollary~\ref{opt_q_statis} since the  performance of loss with respect to the number of rounds is equivalent to that with respect to wall-clock time for homogeneous systems.

\section{Conclusion and Future Work}
\label{sec:conclusion}
In this work, we studied the optimal  client sampling strategy that addresses the system and statistical heterogeneity in FL to minimize the wall-clock convergence time. We obtained a new tractable convergence bound for FL algorithms with arbitrary client sampling probabilities. Based on the bound, we formulated a non-convex wall-clock time minimization problem. We developed an efficient algorithm to learn the unknown parameters in the convergence bound and designed a low-complexity algorithm to approximately solve the non-convex problem. 
Our solution characterizes the interplay between clients' communication delays (system heterogeneity) and data importance (statistical heterogeneity), and their impact on  the  optimal  client sampling design. 
Experimental results validated the superiority of our proposed scheme compared to several baselines in speeding up wall-clock convergence time. 

\clearpage

\bibliographystyle{IEEEtran}
\bibliography{ref}
\end{document}